%% file: 0-main.tex
\documentclass[journal]{ieeeaccess}

\usepackage{cite}

\usepackage{amsmath,amssymb,amsfonts}
\usepackage{bm}
\usepackage{mathrsfs}
\usepackage{color}

\usepackage{graphicx}

\usepackage{textcomp}
\usepackage{calc}
\usepackage{url}
\usepackage{pifont}

\usepackage{array}
\usepackage{tabularx}
\usepackage{booktabs}
\usepackage{multirow}
\usepackage{threeparttable}
\usepackage{etoolbox}
\usepackage[ruled,vlined]{algorithm2e}

\usepackage{listings}

\usepackage{manyfoot}
\usepackage{hhline}

\newcommand{\cmark}{\ding{51}}%
\newcommand{\xmark}{\ding{55}}%

\usepackage{etoolbox}

\let\cline\cmidrule

\newcolumntype{P}[1]{>{\centering\arraybackslash}p{#1}}
\usepackage[detect-none,binary-units=true]{siunitx}[=v2]
\DeclareSIUnit{\nothing}{\relax}
\makeatletter
\AtBeginDocument{\DeclareMathVersion{bold}
\SetSymbolFont{operators}{bold}{T1}{times}{b}{n}
\SetSymbolFont{NewLetters}{bold}{T1}{times}{b}{it}
\SetMathAlphabet{\mathrm}{bold}{T1}{times}{b}{n}
\SetMathAlphabet{\mathit}{bold}{T1}{times}{b}{it}
\SetMathAlphabet{\mathbf}{bold}{T1}{times}{b}{n}
\SetMathAlphabet{\mathtt}{bold}{OT1}{pcr}{b}{n}
\SetSymbolFont{symbols}{bold}{OMS}{cmsy}{b}{n}
\renewcommand\boldmath{\@nomath\boldmath\mathversion{bold}}
\DeclareRobustCommand{\bf}{\bfseries}
\DeclareRobustCommand{\rm}{\rmfamily}
\DeclareRobustCommand{\textbf}[1]{{\bfseries #1}}
\DeclareRobustCommand{\textrm}[1]{{\rmfamily #1}}

}
\makeatother

\begin{document}

\history{Date of publication xxxx 00, 0000, date of current version xxxx 00, 0000.}
\doi{10.1109/XXXX}
\title{MR2-ByteTrack: CNN and Transformer-based Video Object Detection for AI-augmented Embedded Vision Sensor Nodes}

\author{
Luca~Bompani\authorrefmark{1},~Manuele~Rusci\authorrefmark{2},~Luca~Benini\authorrefmark{1}\authorrefmark{3},~Daniele~Palossi\authorrefmark{3}\authorrefmark{4},~Francesco~Conti\authorrefmark{1}%
}
\address[1]{Electrical, Electronic and Information Engineering (DEI), University of Bologna, Italy.}%
\address[2]{Department of Electrical Engineering (ESAT), KU Leuven, Belgium.}%
\address[3]{Integrated Systems Laboratory (IIS), ETH Z\"{u}rich, Switzerland.}%
\address[4]{Dalle Molle Institute for Artificial Intelligence (IDSIA), USI--SUPSI, Switzerland.}%
\tfootnote{This work was supported in part by the Horizon Europe High Performance, Safe, Secure, Open-Source Leveraged RISC-V Domain-Specific Ecosystems project grant agreement 101112274}




\begin{abstract}
Modern smart vision sensors need on-device intelligence to process video streams, as cloud computing is often impractical due to bandwidth, latency, and privacy constraints. 
However, these sensory systems typically rely on ultra-low-power microcontrollers (MCUs) with limited memory and compute, making conventional video object detection methods, which require feature storage or multi-frame buffering, unfeasible.
To address this challenge, we introduce Multi-Resolution Rescored ByteTrack (MR2-ByteTrack), a Video Object Detection (VOD) method tailored for MCU-based embedded vision nodes.
MR2-ByteTrack reduces computational cost by alternating between full- and low-resolution inference, while linking detections across frames via ByteTrack and correcting misclassifications through the Rescore algorithm, {{\color{blue}} which applies probability union rules to aggregate detection confidence scores across frames.}
We apply our approach to both a CNN-based detector and a Transformer-based model, demonstrating its generality across architectures with fundamentally different spatial processing.
Experiments on ImageNetVID demonstrate that MR2-ByteTrack maintains accuracy, achieving mAP scores of up to 49.0 for the CNN-based models and 48.7 for the Transformer, while reducing multiply–accumulate operations by as much as 53\% for the CNNs and 32\% for the Transformer.
When deployed on GAP9, an ultra-low-power RISC-V multicore MCU, our method yields up to 55\% energy savings compared to processing only full-resolution images, enabling the first real-time Transformer-based VOD on an MCU-class embedded vision node. Code available at \url{https://github.com/Bomps4/Multi_Resolution_Rescored_ByteTrack/tree/IEEE_Access}
\end{abstract}



\maketitle




\input{1-Intro}
\input{2-Related}

\input{3-Method}
\input{4-Transformers}
\input{5-Deployment}
\input{6-Results}
\input{7-Conclusion}

\section*{Data Availability Statement}
Data used in this manuscript are available at the following links: \url{https://cocodataset.org/#home} for the COCO~\cite{cocodataset} dataset used to train our network, ImagenetVID~\cite{ILSVRC15} \url{https://image-net.org/challenges/LSVRC/2017/#vid} used for all the evaluations performed in this work. 
\section*{Acknowledgements}
Generative AI was used throughout the manuscript for grammar and spelling checks.
\bibliography{main}
\vspace{-1cm}
\begin{IEEEbiography}[{\includegraphics[width=1in,height=1.25in,clip,keepaspectratio]{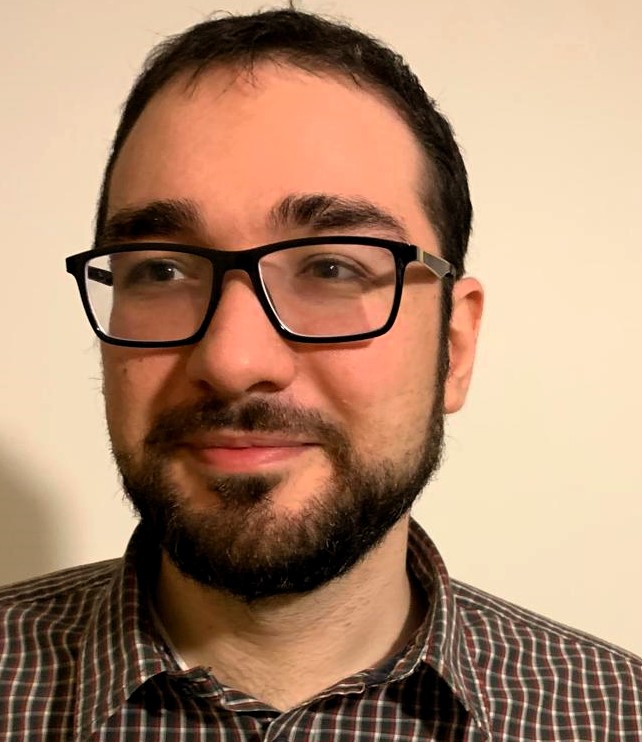}}]{Luca Bompani}
Ph.D. graduate in Electronic Engineering at the University of Bologna, currently a Postdoc researcher at the same institute. He received his MSc in theoretical physics and artificial intelligence at the University of Bologna. His research focuses on deploying and optimizing the inference of Deep Neural Networks for energy-constrained ultra-low-power embedded systems. He received the Best Paper Award at the  CVPR'24 Embedded Vision workshop.
\end{IEEEbiography}
\vspace{-1cm}
\begin{IEEEbiography}[{\includegraphics[width=1in,height=1.25in,clip,keepaspectratio]{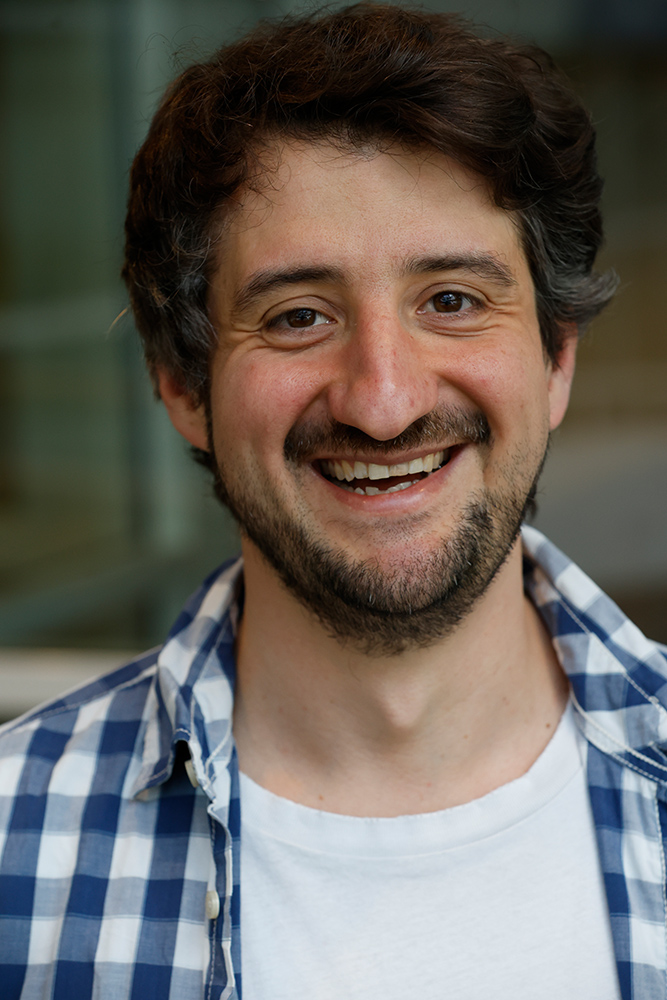}}]{Manuele Rusci}
is currently a tenure-track Assistant Professor at KU Leuven. He received his Ph.D. degree in electronic engineering from the University of Bologna, Italy, in 2018. He has been a postdoctoral fellow at the same university and a Marie Sk\l{}odowska-Curie Actions Postdoctoral grantee at KU Leuven. His main research interests include low-power, AI-powered smart sensors and on-device continual learning.
\end{IEEEbiography}
\vspace{-1cm}
\begin{IEEEbiography}[{\includegraphics[width=1in,height=1.25in,clip,keepaspectratio]{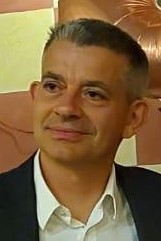}}]{Luca Benini}holds the chair of digital Circuits and systems at ETHZ and is a Full Professor at the Università di Bologna. He received a PhD from Stanford University. His research interests are in energy-efficient parallel computing systems, smart sensing micro-systems, and machine learning hardware. He is a Fellow of the ACM, a member of the Academia Europaea, and of the Italian Academy of Engineering and Technology.
\end{IEEEbiography}
\vspace{-1cm}
\begin{IEEEbiography}[{\includegraphics[width=1in,height=1.25in,clip,keepaspectratio]{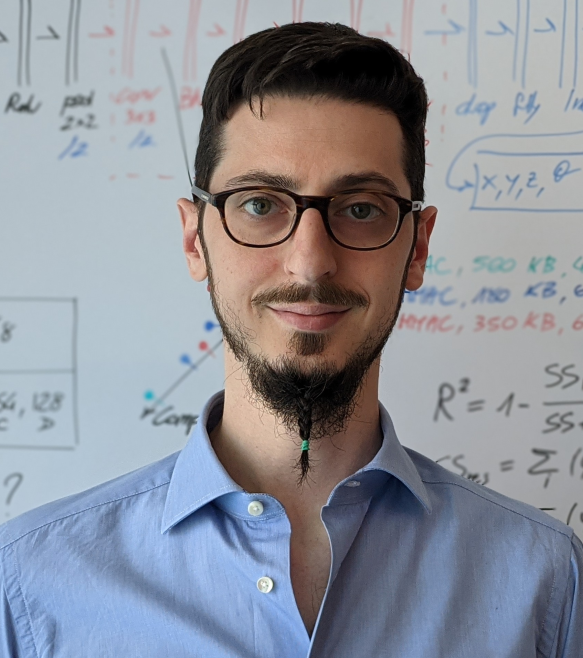}}]{Daniele Palossi}
 (Senior Member, IEEE) received his Ph.D. in Information Technology and Electrical Engineering from ETH Z\"urich. He is currently a Senior Researcher and Lecturer at the Dalle Molle Institute for Artificial Intelligence (IDSIA), USI-SUPSI, Lugano, Switzerland, where he leads the nano-robotics research group, and a Scientific Assistant at the Integrated Systems Laboratory (IIS), ETH Z\"urich, Z\"urich, Switzerland. His research stands at the intersection of artificial intelligence, ultra-low-power embedded systems, and miniaturized robotics. His work has resulted in 50+ peer-reviewed publications in international conferences and journals. Dr. Palossi was a recipient of multiple grants from the Swiss National Science Foundation (SNSF); he received the 2nd prize at the Design Contest held at the ACM/IEEE ISLPED'19, several Best Paper Awards, and led the winning team of the first "Nanocopter AI Challenge" hosted at the IMAV'22 International Conference.
\end{IEEEbiography}
\vspace{-1cm}
\begin{IEEEbiography}
[{\includegraphics[width=1in,height=1.25in,clip,keepaspectratio]{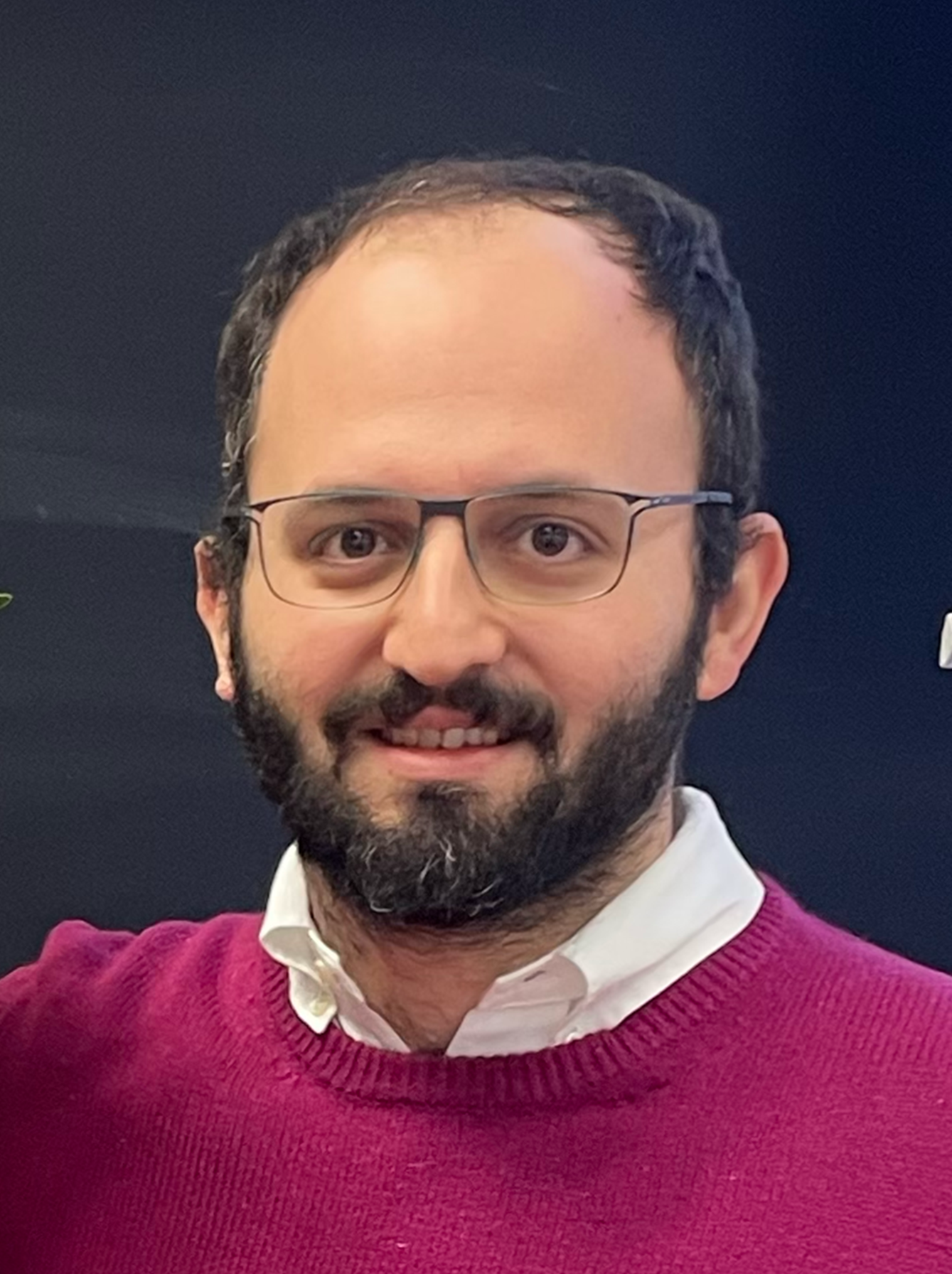}}]{Francesco Conti}
(Senior Member, IEEE) received his Ph.D. degree in electronic engineering from the University of Bologna, Italy, in 2016. He is currently an Associate Professor with the DEI Department at the University of Bologna. From 2016 to 2020, he held a research grant with the University of Bologna and a position as a Post-Doctoral Researcher with ETH Z\"urich. His research is centered on hardware acceleration in ultra-low-power and highly energy-efficient platforms, with a particular focus on System-on-Chips for Artificial Intelligence applications. His research work has resulted in more than 120 publications in international conferences and journals and was awarded several times, including the 2020 IEEE \textsc{Transactions on Circuits and Systems I: Regular Papers} Darlington Best Paper Award.
\end{IEEEbiography}
\EOD
\end{document}

%% file: 1-Intro.tex
\section{Introduction}\label{sec:intro}

Artificial intelligence–based embedded vision is rapidly reshaping the landscape of modern sensing systems, as billions of AI-augmented sensors proliferate across civil, industrial, and consumer applications~\cite{AI_Sensors}.
As these systems scale, it becomes increasingly necessary for sensors to~\textit{process and interpret data directly at the edge}~\cite{AI_on_edge}, while relying on cloud-based computation is often impractical~\cite{Edge_compuyting_survey} due to bandwidth limitations, latency constraints, and the privacy risks of continuous data transmission.

Bringing intelligence to the edge by integrating AI-based algorithms on edge sensor platforms has, therefore, become a key enabler of next-generation autonomous and resource-efficient sensing platforms. Within this broader context, our work focuses on \textit{embedded vision sensors}, which denote a sensing node that integrates an image sensor (within a camera module) with onboard compute and memory, enabling visual information to be partially or fully processed locally on the node.
These nodes must be able to perform various vision tasks, such as object detection, localization, and tracking, which form the backbone of numerous embedded vision applications, including crop health monitoring~\cite{Agriculture}, structural health monitoring~\cite{Structural_health}, wearable assistance~\cite{werableAI}, and mobile robotics~\cite{lamberti2023bio,alnuaimi2022deep}.
Integrating such capabilities directly into sensor nodes, however, introduces significant challenges.
Embedded sensors must typically operate within tight power budgets (a few hundreds of \si{\milli\watt}) and are built around Microcontroller Units (MCUs) with only a few \SI{}{\mega\byte} of memory and sub-\SI{100}{\giga Ops/\second} compute capability~\cite{VegaDSP}.

To operate within their tight computational and memory budgets, embedded vision sensors have traditionally relied on lightweight, frame-by-frame Deep Neural Network (DNN)-based object detectors~\cite{lamberti2023bio,lamberti2021low,alnuaimi2022deep}.
While computationally efficient, these approaches neglect temporal information and therefore sacrifice accuracy.
Video Object Detection (VOD) pipelines address this limitation by exploiting temporal information, obtained by jointly processing several consecutive frames.
Analyzing short sequences enables the model to exploit temporal consistency, as objects in videos tend to move smoothly and remain present over time, thereby stabilizing detections under occlusions, motion blur, or changes in appearance.
However, these conventional VOD methods must process large multi-frame feature maps, making their computational and memory requirements prohibitive for MCU-class devices.

We address this limitation by enabling efficient VOD execution on MCU-equipped edge sensor nodes through a novel \emph{Multi-Resolution Rescored ByteTrack (MR2-ByteTrack)} methodology for deep learning models deployed on sensory systems.
MR2-ByteTrack employs a multi-resolution inference strategy to reduce computational cost by alternating one full-resolution (full-res) inference with several low-resolution (low-res) inferences, thereby lowering the average computational load.
To mitigate accuracy losses caused by motion blur, suboptimal lighting, occlusions, and varying input scales, a lightweight Kalman filter is used to link detections across frames and estimate object motion.
{{\color{blue}}Additionally, MR2-ByteTrack incorporates a mechanism, called the Rescore algorithm, to correct misclassified objects by updating detector confidence scores across frames based on the agreement or disagreement between tracker and detector predictions: 
 accumulating confidence in the former case and penalizing confidence proportionally to the conflicting evidence in the latter}, effectively transforming a frame-by-frame object detector into a full-fledged VOD pipeline.

In our previous work~\cite{ours_conf}, we introduced the MR2-ByteTrack methodology on CNN-based detectors. Here, we extend the precedent study to include Vision Transformers (ViTs)~\cite{ViT}, which have emerged as a powerful alternative to CNNs by capturing global, semantic information and often achieving superior performance in complex visual environments~\cite{Trans_vs_cnn,Trans_vs_cnn2}. Their deployment on ultra-low-power MCUs, however, is more challenging, as ViT-based detectors impose substantial computational demands, with the all-to-all global attention mechanism being the most significant contributor. Despite the architectural differences between CNNs and Transformers, we demonstrate that MR2-ByteTrack can be applied to both architectures, yielding efficiency gains regardless of their distinct spatial computation mechanisms. In particular, our contributions in this paper are as follows:

\begin{itemize}
    \item We present the MR2-ByteTrack methodology, detailing its implementation and providing insight into its application to both CNNs and ViT-based architectures.

    \item We apply our MR2-ByteTrack methodology to two CNN detectors and a hybrid detector, which combines a State-of-the-Art (SotA) Transformer backbone, EfficientViT~\cite{EfficientViT}, with a lightweight CNN head. 

    \item We deploy, to the best of our knowledge, for the first time, a Transformer-based object detector, both as a standalone model and within a complete multi-resolution VOD pipeline, an ultra-low-power MCU, namely GAP9 by Greenwaves Technology (GWT). We provide a detailed analysis of throughput, energy consumption, and resource utilization.
\end{itemize}

We validate our proposed methodology on the ImagenetVID dataset~\cite{ILSVRC15}, demonstrating that MR2-ByteTrack reduces computational cost by 53\% for Shufflenetv2-NanoDet-Plus~\cite{nanodet}(NanoDet) and 32\% for both CSPDarknet-YOLOX-Nano~\cite{ge2021yolox}(YOLOX) and the hybrid detector EfficientViT-B0-YOLOX-Nano (EffVIT-Det)  while preserving their mean Average Precision (mAP).
These results confirm that our multi-resolution strategy, combined with the \textit{Rescore} algorithm, maintains accuracy while significantly reducing computational requirements.
Beyond improving the efficiency of deep visual models, our work advances AI-augmented sensory systems by bringing ViT-level representational capacity to VOD on resource-constrained embedded sensor nodes.
This enables embedded video sensors to process continuous video streams and make reliable decisions under tight computational and energy constraints, thereby reducing or eliminating outright the need for external computation and supporting real-time, adaptive, and resource-aware operation.

%% file: 2-Related.tex
\section{Related Works}\label{sec:related}

\begin{table*}[t]
\caption{Comparison of SotA VOD systems. Each method's improvements (if any, otherwise marked as --) are given w.r.t. its baseline model, keeping the same computational device, e.g., YOLOX-S vs. YOLOV-S in row six.}
\newcolumntype{C}{>{\centering\arraybackslash}X}
\def\arraystretch{1.2}
\begin{tabularx}{\textwidth}{p{4cm} *{6}{C}}
\hline
\multirow{2}{*}{\textbf{Method}} & \multicolumn{2}{c}{\textbf{Improvement}} & \textbf{Detections} & \textbf{Fine-tuning} & \textbf{MCU} & \textbf{Transformer} \\
& \textbf{mAP} & \textbf{Frame/s} & \textbf{aggregation} & \textbf{free} & \textbf{deployable} & \textbf{backbone}\\ 
\hline
3D Convolutions ~\cite{3D_V_CONV} & +7.5 & -- & \xmark & \xmark & \xmark & \xmark \\
FGFA ~\cite{FGFA} & +3.4 & -- & \xmark & \xmark & \xmark & \xmark\\ 
SELSA ~\cite{Selsa} & +6.6 & -- & \xmark & \xmark & \xmark & \xmark\\ 
MEGA~\cite{MEGA} & +7.5 & -- & \xmark & \xmark & \xmark & \cmark\\ 
TransVOD Lite~\cite{TransvodLite} & +4.5 & +5 & \xmark & \xmark & \xmark & \cmark\\ 
YOLOV~\cite{2023yolov} & +7.8 & -- & \xmark & \xmark & \xmark  & \cmark\\ 
Seq-Bbox~\cite{Seq-BBox} & +6.9 & -- & \cmark & \cmark & \xmark & \xmark\\ 
Motion-based Seq-Bbox~\cite{motion_based} & +5.5 & -- &\cmark & \cmark & \xmark & \xmark\\
Objects Do Not Disappear~\cite{Objectsdonotdisappear} & +13.6 & +30 & \xmark & \xmark & \xmark & \cmark \\ 
Block Copy~\cite{Block_copy} & -- &+3 & \xmark & \xmark & \xmark & \xmark\\
PASS~\cite{Pass} & -- & +9 & \xmark & \xmark & \xmark & \xmark\\
Looking Fast and Slow~\cite{Liu2019LookingFA} & +3.4 & +68 & \xmark & \xmark & \xmark & \xmark\\ 
Frame-by-frame~\cite{lamberti2023bio,Dsortmcu} & -- & -- & \xmark & \xmark & \cmark & \xmark\\ 
\textbf{Our (NanoDet)} & \textbf{+7.3} & \textbf{+4} & \cmark & \cmark & \cmark & \xmark\\ 
\textbf{Our (YOLOX)} & \textbf{+5.3} & \textbf{+1.5} & \cmark & \cmark & \cmark & \xmark\\ 
\textbf{Our (EffVIT-Det)} & \textbf{+3.8} & \textbf{+1.2} & \cmark & \cmark & \cmark & \cmark\\ 
\hline
\end{tabularx}

\label{table:vod_table}
\end{table*}

DNN-based video object detectors extend typical object detection models for the processing of video sequences. 
Simple VOD strategies, which are commonly used for resource-constrained embedded sensor systems, simply run object detection models frame by frame.
On the other hand, more complex methods can aggregate information from multiple frames to determine the bounding boxes, class labels, and object confidence scores in a video.


{{\color{blue}}

The design of VOD systems for MCU-class devices is driven by two hard constraints: available memory, typically a few MB, and the strict causal and real-time requirements of operations.
By causal, we mean that each per-frame inference may rely only on information from the current frame and those preceding it, never on future frames.
Most existing VOD approaches violate one or both constraints: they either require storing large intermediate feature tensors across frames~\cite{Block_copy,Objectsdonotdisappear,TransvodLite,3D_V_CONV,FGFA}, depend on multi-stage architectures with variable memory consumption~\cite{MEGA,Selsa}, or rely on future frames to refine past detections or improve efficiency~\cite{seqnms,Seq-BBox,motion_based}.
The following review is organized around these two axes, highlighting at each step where prior approaches fall short for MCU-class platforms and the specific gaps that MR2-ByteTrack addresses.
}

\subsection{Lightweight Object Detection}\label{subsec:detectors}

Object detection methods can be categorized into two-stage and one-stage approaches.
Two-stage models, such as Faster R-CNN~\cite{Faster_RCNN}, iterate over the region proposals produced by the first stage to identify objects in an image. 
Given the generally high number of proposals, i.e., regions of the image where the network predicts there can be an object, these detectors are typically computationally expensive and unsuitable for power-constrained devices.
On the other hand, one-stage detectors, including SSD~\cite{SSD} and its optimized version SSDLite~\cite{MobileNetv2}, directly predict object locations and categories in a single pass, thereby reducing complexity and improving efficiency compared to two-stage methods. 
More specifically, SSDLite features \SI{4.3}{\mega\nothing} parameters and achieves a mAP of 22.1 on COCO, while requiring \SI{0.8}{\giga MAC} operations.
This work focuses, therefore, on one-stage detectors due to their more compact architecture and predictable execution time.

An active research line aims to reduce the computation and memory requirements of object detection models for deployment on resource-constrained devices. 
Among the notable examples, NanoDet~\cite{nanodet} features \SI{1.17}{\mega\nothing} parameters and reaches an mAP of 27 on COCO by executing only \SI{0.46}{\giga MAC} operations.
The unprecedented detection accuracy is achieved thanks to two novel auxiliary layers that improve the training process, namely the Assign Guidance and the Dynamic Soft Label Assigner.
YOLOX~\cite{ge2021yolox}, which is a lightweight architecture derived from YOLOv3, introduces depthwise convolutions and an anchor-free decoupled head. 
This network scores a mAP of 25.8 with \SI{0.36}{\giga MAC} operations and \SI{0.9}{\mega\nothing} parameters.
In contrast, more recent YOLO variants, such as YOLOv8~\cite{YOLOv8}, YOLOv10~\cite{YOLOv10}, and YOLOv11~\cite{YOLOv11}, focus on balancing detection accuracy and computational efficiency but still require significant compute resources (more than \SI{6.5}{\giga MAC} per inference).
In this work, we leverage efficient architectures, such as NanoDet and YOLOX, within the components of our VOD pipeline. 

Recent works have also described object detection algorithms based on Transformers. 
DETR~\cite{DETR} is one of the first instances of this class of method, which made use of a new bipartite matching loss for efficient training. 
Unfortunately, its high memory footprint (\SI{159}{\mega \byte}) prevents the deployment on constrained platforms. 
In contrast, mobile-friendly Transformer architectures like MobileViT~\cite{mobilevit, mobilevit3, mobilevit2} achieve an mAP of 19.3 with only \SI{1.2}{\mega\nothing} parameters for the backbone.

Because this work targets ultra-efficient model deployment ($<$\SI{1}{\mega\nothing} parameters for our backbones), we propose a new transformer-based object detector based on EfficientViT-B0~\cite{EfficientViT}.
This backbone features only \SI{0.7}{\mega\nothing} parameters but was previously employed only for classification or segmentation tasks.

\subsection{Video Object Detection} \label{subsec:VOD_related}

Typical VOD algorithms extend the object detectors discussed in the previous section to process video streams.
One of the earliest VOD strategies replaces 2D convolution layers in CNNs with 3D convolutions to concurrently process a batch of frames.
This technique aims to model spatiotemporal correlations across features extracted from successive frames~\cite{tran2015learning,3D_V_CONV}.
Despite its effectiveness, this approach requires significant memory to store high-dimensional activation tensors, making this class of methods not portable on memory-constrained platforms. 
Alternative methods accumulate and process the visual features extracted by every frame to infer the temporal information. 

When two-stage detectors are used, feature aggregation is commonly performed by aggregating features relative to region proposals.
The temporal information is then extracted by using techniques such as optical flow~\cite{FGFA}, convolutional-based trackers~\cite{LYU2021139, IntegratedOD}, feature similarity~\cite{Selsa}, or Transformer-based memory layers~\cite{MEGA}. 
These methods ultimately utilize the aggregated information to refine video detections, thereby ensuring temporal consistency. However, the use of a two-stage approach, which cannot be efficiently deployed due to its varying memory consumption and the extra memory required for storing the features, makes this method impractical for our scenario.

In contrast, one-stage object detectors integrate localization and classification into a single pipeline, resulting in diverse solutions for temporal feature aggregation.
TransVOD LITE~\cite{TransvodLite} aggregates full feature maps across frames. 
Starting from a Deformable DETR~\cite{DeformableDD}, this method inserts a Sequential Hard Query Mining layer between the feature extractor and the detection head.
The new layer aggregates features from multiple frames to enhance temporal consistency.
As a drawback, this method requires keeping the feature maps extracted from the video frames in memory.
Given that the authors considered a temporal window of 15 frames, we estimated a memory overhead of \SI{272}{\mega\byte}, which exceeds the memory resource of our devices. 
YOLOV~\cite{2023yolov}, on the other hand, directly refines the detections provided by the underlying object detector outputs based on the features accumulated across frames. While efficient, this method is impractical for MCU due to its non-constant memory consumption, which is dependent on the number of objects found in each frame and cannot be optimized at deployment time.

These approaches are not suitable for our purpose due to the high memory requirements necessary to preserve the extracted features over time.
Furthermore, the presented solutions utilized additional feature aggregation layers that required fine-tuning over video sequences, i.e., computationally and data-expensive extra training. 

A more lightweight approach than temporal feature aggregation directly searches for temporal correlations over the frame-by-frame detections.
The work~\cite{seqnms} utilizes dynamic programming to link objects over time, employing the Intersection over Union (IoU) as a cost function.
Seq-Bbox~\cite{Seq-BBox} and Motion-based Seq-Bbox~\cite{motion_based} aggregate the detection outputs by using a similarity metric.
The latter methods achieve mAP scores of 80.9 and 72.7 on the ImageNetVID dataset, showing respective gains of +6.9 and +5.5 mAP points over frame-by-frame inference.
The "Objects do not disappear" method ~\cite{Objectsdonotdisappear} utilizes the object trajectory as an additional supervision signal during training, achieving an mAP of 91.3.
The proposed block, namely the trajectory prediction network, can also predict future bounding box locations, thereby preventing the need for a more expensive object detector. 

Our approach has two main differences with respect to the last set of works. 
First, the methods above are optimized only for detection accuracy, whereas our pipeline also improves throughput and energy metrics. 
Second, we target strictly online real-time operation, whereas several previous approaches improve accuracy by buffering multiple frames and refining detections retrospectively, i.e., using frames that are future with respect to the output being corrected.
The only exception is ~\cite{Objectsdonotdisappear}, which uses a trajectory prediction network to predict the object's motion.
However, this comes at a high computational cost of $\sim$\SI{1.3}{\giga MAC} per predicted frame.
While this overhead may be justified for large object detectors, it is impractical for our lightweight models, as it is nearly 3$\times$ more expensive than the inference task.

Our MR2-ByteTrack solution follows a similar approach to~\cite{Objectsdonotdisappear}, but replaces the costly neural network-based motion prediction with a lightweight Kalman filter. 
This design choice is inspired by two SoA object tracking methods: Simple Online Realtime Tracking (SORT)~\cite{sort}, which uses the Hungarian algorithm for associating detections with the tracked objects, and ByteTrack~\cite{ByteTrackMT}, which enhances SORT by integrating low-confidence detections to retain object identities even when occlusions and appearance change.
MR2-ByteTrack improves ByteTrack by (i) adapting the approach for a multi-class detection problem while the original work only faces class-agnostic tracking tasks and (ii) introducing Rescore, a probabilistic classification correction method that improves multi-class accuracy.

\subsection{Efficient Video Processing}\label{subsec:VODeff_related}

Unlike previous methods that only target accuracy optimization, temporal information can also be leveraged to enhance computation and memory efficiency in VOD tasks. 
Consecutive frames often contain redundant information, making full feature recomputation not necessary. 
Hence, methods can reuse features or recompute new features only from specific regions of the image. 
For instance, in \cite{Block_copy}, a policy network selectively recomputes only necessary regions, while \cite{Pass} employs a gating mechanism in convolutional layers to decide whether to compute new features or reuse the existing ones.
However, these approaches require storing large activation maps, up to $\sim$\SI{50}{\mega\byte} for NanoDet at 320$\times$320 pixels, which far exceeds MCU memory constraints.

Another efficiency technique is model interleaving, where high- and low-complexity models are alternatively invoked to trade off accuracy and speed. 
Among the others, \textit{Liu et al.}~\cite{Liu2019LookingFA} use an LSTM to associate detections returned by a large (\SI{4.4}{\mega\nothing} parameters) and a small (\SI{0.5}{\mega\nothing} parameters) detectors.
\textit{Moretti et al.}~\cite{Motetti2024AdaptiveDL} proposed to dynamically switches models based on computational constraints. 
Despite their effectiveness, these methods require multiple models or custom architectures, which doubles (or multiplies by a number of models) the deployment costs, e.g., memory for weights or binary code.
Our approach, in contrast, utilizes a single pre-trained model and scales the computational efficiency by varying the input resolution of the data.

\subsection{Object Detection and VOD on Embedded Sensor Nodes} \label{subsec:VOD_MCU}

Existing SoA VOD approaches for MCU-based devices typically perform inference on a frame-by-frame basis, without exploiting temporal correlations between consecutive frames.
\textit{Lamberti et al.}~\cite{lamberti2023bio} demonstrated the deployment of a quantized MobileNetV2-SSD on the GAP8 MCU, achieving only \SI{1}{frame/\second} after 8-bit quantization.
Throughput can be improved by leveraging on-chip accelerators: the authors of~\cite{Dsortmcu} implemented TinyssimoYOLO on the GAP9 MCU, reaching approximately \SI{60}{frames/\second} by exploiting the NE16 8-bit convolution accelerator~\cite{TinyssimoYOLO}.
Despite its efficiency, this model is highly specialized, detecting only a single object class (cars).
A subsequent refinement~\cite{TinyYOLOv2} extended the network to multiple classes and achieved \SI{117}{frames/\second} at a 112$\times$112 input resolution; however, in the evaluation, this network achieves only 15  mAP on the COCO dataset, 10.8 less than the YOLOX nano used in our work with a similar parameter count.

More recently, StreamTinyNet~\cite{Streamtiny} was introduced for video streaming analysis. 
Although not explicitly designed for VOD, it uses feature aggregation to reduce memory demands by compressing features from T frames into the size of a single frame using pointwise convolutions. 
However, this approach requires training a new model from scratch, whereas our method works with any pre-trained object detector. Furthermore, it still requires storing the features across video frames, increasing memory consumption.
Finally, our references show that the current SoA for video object detection and object detection on MCUs relies on CNNs, as shown in~\cite{Survey_CNNs_on_MCU}. While Transformers have been deployed on MCUs for classification tasks, e.g., MCUformer and TinyViT, among other approaches~\cite{Deployment_trans,Deploy_transformers,Transformer_mio,McuFormer}, to the best of the authors' knowledge, no Transformer-based methods have yet been used for object detection on MCUs.

{{\color{blue}}Table~\ref{table:vod_table} summarizes the major OD and VOD approaches most relevant to our work and confirms that no existing method satisfies both deployment constraints simultaneously. Feature-level temporal aggregation methods (Section~\ref{subsec:VOD_related}) buffer high-dimensional tensors across frames, incurring memory overheads of tens to hundreds of MB~\cite{MEGA,2023yolov,TransvodLite}; our method operates directly on detection outputs, introducing no feature storage overhead.
Detection-level and tracking-based methods (Section~\ref{subsec:VOD_related}) are lighter but achieve only limited efficiency gains and frequently exploit future frames to refine past detections~\cite{seqnms,Seq-BBox,motion_based}; our pipeline is strictly causal and improves both accuracy and throughput.
Efficiency-oriented approaches based on multi-resolution execution or model interleaving (Section~\ref{subsec:VODeff_related}) either retain large intermediate activations~\cite{Block_copy} or require multiple networks to be simultaneously present in memory~\cite{Liu2019LookingFA}; our method deploys a single pre-trained detector with no additional runtime buffers or fine-tuning, making it directly applicable to any pre-trained object detector.
Existing MCU deployments (Section~\ref{subsec:VOD_MCU}) sidestep these issues by discarding temporal information entirely~\cite{lamberti2021low,lamberti2023bio}; our approach recovers the resulting accuracy and throughput gains without violating either deployment constraint, balancing between accuracy, throughput, energy efficiency, and deployment cost, making it well-suited for low-power, real-time VOD on MCU-class platforms.}

%% file: 3-Method.tex
\begin{figure*}[t]
\includegraphics[width=\textwidth]{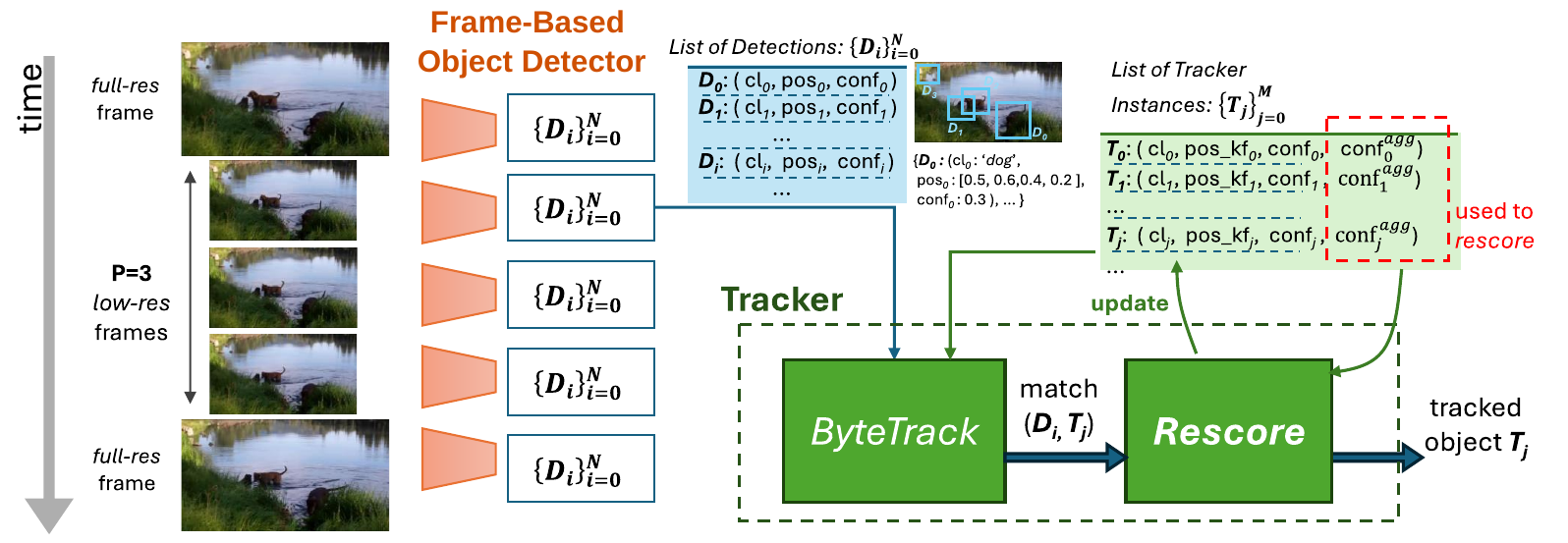}
\caption{Overview of the MR2-ByteTrack algorithm.}
\label{fig:overview}
\end{figure*}

\section{MR2-ByteTrack}\label{MRBytetr}

{{\color{blue}}Figure~\ref{fig:overview} illustrates the main building blocks of MR2-ByteTrack, while Algorithm~\ref{algo:pipeline} details the operations performed at each inference step.}
First, the video frames are processed in real-time using a CNN- or Transformer-based object detection algorithm. 
We adopt a \textit{Multi-Resolution} scheme where the resolution of the input images varies over time. 
Second, a \textit{Tracker} module, derived from ByteTrack \cite{ByteTrackMT}, filters detections based on their motion, and, lastly, a \textit{Rescore} algorithm refines the predicted object classes and confidence scores to address potential mispredictions. 
The following subsections provide more details on these components.

\subsection{Multi-resolution Object Detection}\label{subsec:multires}

MR2-ByteTrack works with convolution- or Transformer-based object detectors. 
The models are fed input images of varying resolutions. 
More specifically, we interleave one full-resolution (full-res) image (e.g., 320$\times$320 px) with P low-resolution (low-res) images (e.g., 192$\times$192 px).
Compared to using only full-res frames, this multi-resolution scheme consistently reduces the computation cost of the detection algorithm, as measured by the total number of required MAC operations: 
\begin{equation}
\label{eq:mac}
\text{MAC} = \rho \text{MAC}_{fr} + (1-\rho)\text{MAC}_{lr}
\end{equation}
where $\rho = \frac{1}{1+P}$, $\mathrm{MAC}^{fr}$ and $\mathrm{MAC}^{lr}$ represent the number of MAC operations required to execute the inference on one full-res and one low-res frame, respectively.
The loss of detection accuracy resulting from interleaving low-res frames versus using only full-res frames is counteracted by utilizing ByteTrack and the Rescore algorithms, as described in Sections \ref{subsect:ByteTrack} and \ref{subsect:Rescore}.
Additionally, we note that the object detectors reuse the same parameters during both full-resolution and low-resolution inferences. 
Hence, by design, the weight memory footprint of our multi-resolution setting matches those of baseline single-resolution detectors.

In the case of CNNs, the reduction of MAC operations when computing with low-res frames is related to the spatial sliding of the convolution filters over the input image and the intermediate results.
Generally,  the total number of MACs scales linearly with respect to the area of the input image, e.g., for the CNN-based NanoDet object detector.

On the other side, Transformers process image patches of fixed sizes (e.g., 16$\times$16 px in ViT), and their attention layers compute the correlation between the features extracted by the patches~\cite{transf_original}. 
Hence, by reducing the resolution of the input image, a Transformer processes a lower number of patches, scaling the total number of operations quadratically with respect to the area of the input data. 
In this work, we apply the multi-resolution scheme to a novel Transformer-based object detector featuring linear attention layers (Sec~\ref{sec:tranforms}).
For this attention mechanism, the computation cost of the inference task scales linearly with respect to the input area, as explained in the following section.


\RestyleAlgo{ruled}
\begin{algorithm}[]
{\footnotesize
$\mathbf{Inputs:}$ Frame $F^t$, active trackers $\{T_j\}_{j=1}^{M}$ \\
\quad $F^t$: RGB image frame at time $t$ \\
\quad $\{T_j\}_{j=1}^{M}$: set of active trackers, $T_j = (\mathrm{bbox}_j^{\mathrm{kf}},\, \mathrm{cl}_j,\, \mathrm{conf}_j)$ \\
\quad \quad $\mathrm{bbox}_j^{\mathrm{kf}}$: bounding box in image coordinates, $\mathrm{cl}_j$: class index, $\mathrm{conf}_j$: confidence score \\
\BlankLine
$\mathbf{Model\ Parameters:}$ $\mathit{high\_threshold}$, $\mathit{low\_threshold}$, $\tau_{\mathrm{IoU}}$, $\tau_{\mathrm{init}}$, $\tau_{\mathrm{dead}}$, \text{P} \\
\BlankLine

\eIf{$t \bmod (P+1) \neq 0$}
{
    $\hat{F}^t \gets \mathrm{Resize}(F^t)$ \tcp*{low-resolution frame}
}
{
    $\hat{F}^t \gets F^t$ \tcp*{full-resolution frame}
}
\BlankLine

$\{D_i\}_{i=1}^{N} \gets \mathrm{ObjectDetector}(\hat{F}^t)$,\quad $D_i = (\mathrm{bbox}_i,\, \mathrm{cl}_i,\, \mathrm{conf}_i)$ \\
\BlankLine

$\{D_i^{\mathrm{high}}\} \gets \{D_i \mid \mathrm{conf}_i \geq \mathit{high\_threshold}\}$ \\
$\{D_i^{\mathrm{rem}}\} \gets \{D_i \mid \mathit{low\_threshold} \leq \mathrm{conf}_i < \mathit{high\_threshold}\}$ \\
\BlankLine

\tcp{First association pass: high-confidence detections vs. all active trackers}
$\mathbf{C}^{\mathrm{high}} \gets \mathrm{IoUMatrix}(\{D_i^{\mathrm{high}}\},\, \{T_j\}_{j=1}^{M})$ \\
$(\mathcal{M}^{\mathrm{high}},\, \mathcal{U}_D^{\mathrm{high}},\, \mathcal{U}_T^{\mathrm{high}}) \gets \mathrm{Match}(\mathbf{C}^{\mathrm{high}},\, \tau_{\mathrm{IoU}})$ \\
\ForEach{$(i,j) \in \mathcal{M}^{\mathrm{high}}$}{
    $T_j \gets \mathrm{KalmanUpdate}(T_j,\, D_i)$
}
\ForEach{$i \in \mathcal{U}_D^{\mathrm{high}}$}{
    initialize new tracker from $D_i$ if confirmed after $\tau_{\mathrm{init}}$ consecutive matches
}
\BlankLine

\tcp{Second association pass: $\{D_i^{\mathrm{rem}}\}$ vs. unmatched trackers only}
$\mathbf{C}^{\mathrm{rem}} \gets \mathrm{IoUMatrix}(\{D_i^{\mathrm{rem}}\},\, \{T_j \mid j \in \mathcal{U}_T^{\mathrm{high}}\})$ \\
$(\mathcal{M}^{\mathrm{rem}},\, \_,\, \_) \gets \mathrm{Match}(\mathbf{C}^{\mathrm{rem}},\, \tau_{\mathrm{IoU}})$ \\
\ForEach{$(i,j) \in \mathcal{M}^{\mathrm{rem}}$}{
    $T_j \gets \mathrm{KalmanUpdate}(T_j,\, D_i)$
}
\tcp{unmatched $D_i^{\mathrm{rem}}$ are discarded}
\BlankLine

\tcp{Propagate unmatched trackers via Kalman prediction; remove stale ones}
\ForEach{$T_j$ with $j \in \mathcal{U}_T^{\mathrm{high}}$ unmatched after both passes}{
    $T_j \gets \mathrm{KalmanPredict}(T_j)$ \\
    \If{$T_j$ unmatched for $\tau_{\mathrm{dead}}$ consecutive steps}{remove $T_j$}
}
\BlankLine

$\{T_j\}_{j=1}^{M} \gets \mathrm{Rescore}\!\left(\{T_j\}_{j=1}^{M}\right)$ \\
\BlankLine
\Return $\{T_j\}_{j=1}^{M}$
}
\caption{MR2-ByteTrack tracking pipeline.}
\label{algo:pipeline}
\end{algorithm}

\subsection{ByteTrack}\label{subsect:ByteTrack}

To recover from potential misdetections caused by low-resolution frames, a Kalman-filter-based tracking algorithm, ByteTrack, is applied to the detections produced at each frame.
Formally, given a frame F at time t, the object detector outputs a set of detections $\{D_i\}_{i=1}^N$, where N is the total number of detected objects in the current frame.
As can be seen from Figure~\ref{fig:overview}, each detection $D_i$ is represented by a triplet ($\text{bbox}_i$, $\text{cl}_i$, $\text{conf}_i$), where: $\text{bbox}_i$ is the bounding box defining the object's position in the image, $\text{cl}_i$ is the class index of the object, and $\text{conf}_i$ is the confidence score ($\text{conf}_i<=1$) for the class $\text{cl}_i$.

At each time step t, ByteTrack keeps a list of active trackers $\{T_j\}_{j=1}^M$, where each instance corresponds to a tracked object.
Similarly to detections, each tracker ($T_j$) is represented as a triplet ($\text{bbox}^{kf}_j$, $\text{cl}^{kf}_j$, $\text{conf}^{kf}_j$).
In this triplet, $\text{bbox}^{kf}_j$ is the object's bounding box, which is refined by a Kalman filter based on motion, $\text{cl}_j$ represents the class index assigned to the tracked object, and $\text{conf}_j$ is the confidence score of the assigned class.

To find correspondences between detections $D_{i=1}^N$ and existing trackers $T_{j=1}^M$, MR2-ByteTrack uses the Intersection over Union as a cost function.
A detection is associated with an existing tracked object only if its IoU exceeds the threshold value {{\color{blue}}($\tau_{\mathrm{IoU}}$)} of 0.3.
Unlike the original ByteTrack work, we do not consider feature-matching cost components for associating detections across frames because of the multi-resolution scheme. 
Additionally, a tracker is considered active after matching with two consecutive detections {{\color{blue}}($\tau_{\mathrm{init}}$)}, and is removed if it is not updated during the latest $5$ time steps {{\color{blue}}($\tau_{\mathrm{dead}}$)}. 

Finally, we introduce a filtering mechanism to ensure reliable tracking. 
When a new detection does not match any active tracker, it initializes a new tracker only if its confidence score exceeds a predefined \textit{high\_threshold}. 
Conversely, detections with a confidence score below the \textit{low\_threshold} are discarded. Finally, a detection whose confidence is between the \textit{high\_threshold} and \textit{low\_threshold} is kept only if it can be linked to a previous 
These thresholds are tuned to control the precision (mainly impacted by \textit{high\_threshold}) and the recall (\textit{low\_threshold}) of our algorithm. { {\color{blue}}  Algorithm~\ref{algo:pipeline}, summarizes the pipeline described in this section.}

\subsection{Rescore Algorithm}\label{subsect:Rescore}

The new Rescore algorithm (Algorithm~\ref{algo}) revisits (rescores) the class indexes and the confidence scores assigned by ByteTrack to the tracked objects.
We exploit the temporal consistency of moving objects in a video, i.e., the history of the detections, to identify and correct algorithm misclassification.

For every $T_j$ tracker, we define an additional class status variable ($ \text{conf}^{agg}_i$) to aggregate the confidence scores of all detections matched with $T_j$ up until time step $\text{t}-1$.
When a new detection $D_i$ is assigned to $T_j$ at time t, and they belong to the same class, we update $\text{conf}^{agg}_j$ as: 

\begin{equation} 
\text{conf}^{agg}_j = 1 - ((1 - \text{conf}^{agg}_j)*(1-\text{conf}_i))
\end{equation}
{{\color{blue}} In this update rule, the confidence score $\text{conf}_i$ is interpreted as the probability that the class is correct.
The formula is the complement of the probability that both detections are simultaneously incorrect, i.e., it captures all cases in which at least one of the two detections is correct (including the case in which both are correct).
Therefore, the aggregation increases overall confidence by excluding individual error cases and retaining only the less probable joint-error case.}
If the class $\text{cl}_i$ of the new detection differs from the tracker’s class $\text{cl}_j$, $\text{conf}^{agg}_j$ is reduced as:
\begin{equation} 
\text{conf}^{agg}_j = 1 - ((1 - \text{conf}^{agg}_j)/(1-\text{conf}_i))
\end{equation}
{{\color{blue}}the update rule in this instance represents a normalized confidence margin, where the residual uncertainty of the class carried over by the tracker $(1-\text{conf}^{agg}_i)$ is weighted by the confidence that the competing class is wrong $(1-\text{conf}_i)$.
This reduces the confidence of discordant detections and retains only the portion of confidence that supports the preferred class over its alternative.}
If the confidence score of the new detection ($\text{conf}_i$) exceeds the aggregated score $\text{conf}^{agg}_j$, we modify the tracker’s class to $\text{cl}_i$.
Because any rescore operation can not occur if $\text{conf}^{agg}_j = 1$, the value of the aggregated score is capped after the update to $1-\epsilon$, where $\epsilon$ is $10^{-4}$. 
Finally, Rescore assigns a new confidence score to a tracked object as the mean confidence score of the matched detections.

\RestyleAlgo{ruled}
\begin{algorithm}[t]

{\footnotesize
$\mathrm{\mathbf{Inputs: }}$ Match $ \{ D_i = (\mathrm{cl}_i, \mathrm{conf}_i), T_j = (\mathrm{cl}_j, \mathrm{conf}_j, \mathrm{conf_j^{agg}}) \}$ \\
\eIf {$\mathrm{cl}_j$ == $\mathrm{cl_i}$ }
{$\mathrm{conf_j^{agg}} \gets 1-((1-\mathrm{conf}_i)*(1-\mathrm{conf_j^{agg}}))$
}
{\eIf{$\mathrm{conf_j^{agg}} < \mathrm{conf}_i$}
{
$(\mathrm{cl}_j, \mathrm{conf}_j, \mathrm{conf^{agg}}_j ) \gets (\mathrm{cl}_i, \mathrm{conf}_i, \mathrm{conf}_i)$\\
}
{
$ \mathrm{conf_j^{agg}} \gets 1-((1-\mathrm{conf_j^{agg}})/(1-\mathrm{conf}_i))$\\
$\mathrm{conf_j^{agg}} \gets max(\mathrm{conf_j^{agg}},0)$ \\
\If{$\mathrm{conf_j^{agg}} < \mathrm{conf}_i$}
{
$(\mathrm{cl}_j, \mathrm{conf}_j, \mathrm{conf_j^{agg}} ) \gets (\mathrm{cl}_i, \mathrm{conf}_i, \mathrm{conf}_i)$
}
}
}
$\mathrm{conf_j} \gets \mathrm{mean}\left(\mathrm{conf}^{t}_i\mathrm{conf}^{t-1}_i,\mathrm{conf}^{t-2}_i\right)$
$\mathrm{conf_j^{agg}} \gets min(\mathrm{conf_j^{agg}},1-\epsilon)$\\
\Return $T_j = (\mathrm{cl}_j, \mathrm{conf}_j, \mathrm{conf_j^{agg}}) $
}
\caption{Rescore algorithm.}
\label{algo}
\end{algorithm}

%% file: 4-Transformers.tex
\section{Transformers-based MR2-ByteTrack} \label{sec:tranforms}

Given the lack of lightweight Transformer-based object detectors, we propose a novel architecture named EffViT-Det. 
Our design utilizes the EfficientViT-B0~\cite{EfficientViT} backbone, consisting of 5 convolutional layers followed by three attention-based transformer layers. 
The latter feature two main differences with respect to the layers of traditional Vision Transformers. 
First, the standard softmax-based attention mechanism is replaced with a ReLU-based linear attention mechanism. 
Second, the size of the output tensors of the attention layers decreases as the model progresses toward its output. 
Conversely, ViT presents a fixed feature map size throughout the entire network. 

In our object detector, the hierarchical feature maps produced by the EfficientViT-B0~\cite{EfficientViT} backbone are firstly aggregated and then passed to a detection head, following the design approach previously proposed by YOLOX~\cite{ge2021yolox}. 
For feature aggregation, we utilize a Path Aggregation Network (PAN)~\cite{ge2021yolox}, which aims to fuse high-resolution features with low-resolution semantic features in a bottom-up manner. 
This solution leads to a higher mAP on COCO than using a top-down Feature Pyramid Network (FPN)~\cite{Retina} (24.6 mAP for PAN vs. 22.0 mAP for FPN), while being comparable in terms of computation cost (\SI{281}{\mega MAC} vs. \SI{272}{\mega MAC}). 

When integrated into MR2-ByteTrack, we feed our EffViT-Det object detector with full-resolution or low-resolution images to trade off accuracy versus computation load (Equation~\ref{eq:mac}).
The attention layers of the EfficientViT-B0 backbones process the features extracted by patches of size 8$\times$8 px (first attention layer), 16$\times$16 px (second), or 32$\times$32 px (third).
Following the discussion in~\cite{EfficientViT}, the attention computes the all-to-all similarity across the patches.
Given a total of $n$ patches, the attention between the $i$ and $j$-th blocks, indicated as $A_{ij}$ is computed as:
\begin{equation}\label{eq:onlyattention}
	A_{ij} = \frac{Sim\left( \frac{1}{\sqrt{d_k}} \sum_{k=1}^{d_k} Q_{ik} K_{jk} \right)}
	{\sum_{j'=1}^{n} Sim\left( \frac{1}{\sqrt{d_k}} \sum_{k=1}^{d_k} Q_{ik} K_{j'k} \right)}
\end{equation}
where $i$ and $j$ range from 1 to $n$,  \( d_k \) is the  feature dimension and  $Q_{ik}$ and $K_{ik}$ are the element of, respectively, $d_k$-sized query and key vectors extracted from the i-th patch. 
For our linear attention layers, $Sim()=ReLU()$, hence Equation~\ref{eq:onlyattention} can be rewritten as: 
\begin{equation}\label{eq:lin_attention}
	 A_{ij} = 
    \sum_{k=1}^{d_k} \frac{ \text{ReLU}\left( Q_{ik}\right) \text{ReLU}\left(K_{jk} \right)}
	{\sum_{j'=1}^{n}\text{ReLU}\left( Q_{ik}\right) \text{ReLU}\left(K_{j'k} \right)}
\end{equation}
By simplifying $\text{ReLU}\left(Q_{ik}\right)$, the attention equation no longer depends on the i-th element.
As a result, the number of operations in our EffViT-Det scales linearly with respect to the number of patches, i.e., the resolution of the input image. 
In our MR2-ByteTrack solution, the total number of MAC for transformer-based inference is, therefore, 2.8$\times$ lower when scaling the resolution of the input image from 320$\times$320 px to 192$\times$192 px.

%% file: 5-Deployment.tex
\section{VOD for Embedded Sensor Nodes}\label{subsec:mcu}
\begin{figure}[t]
\includegraphics[width=\columnwidth]{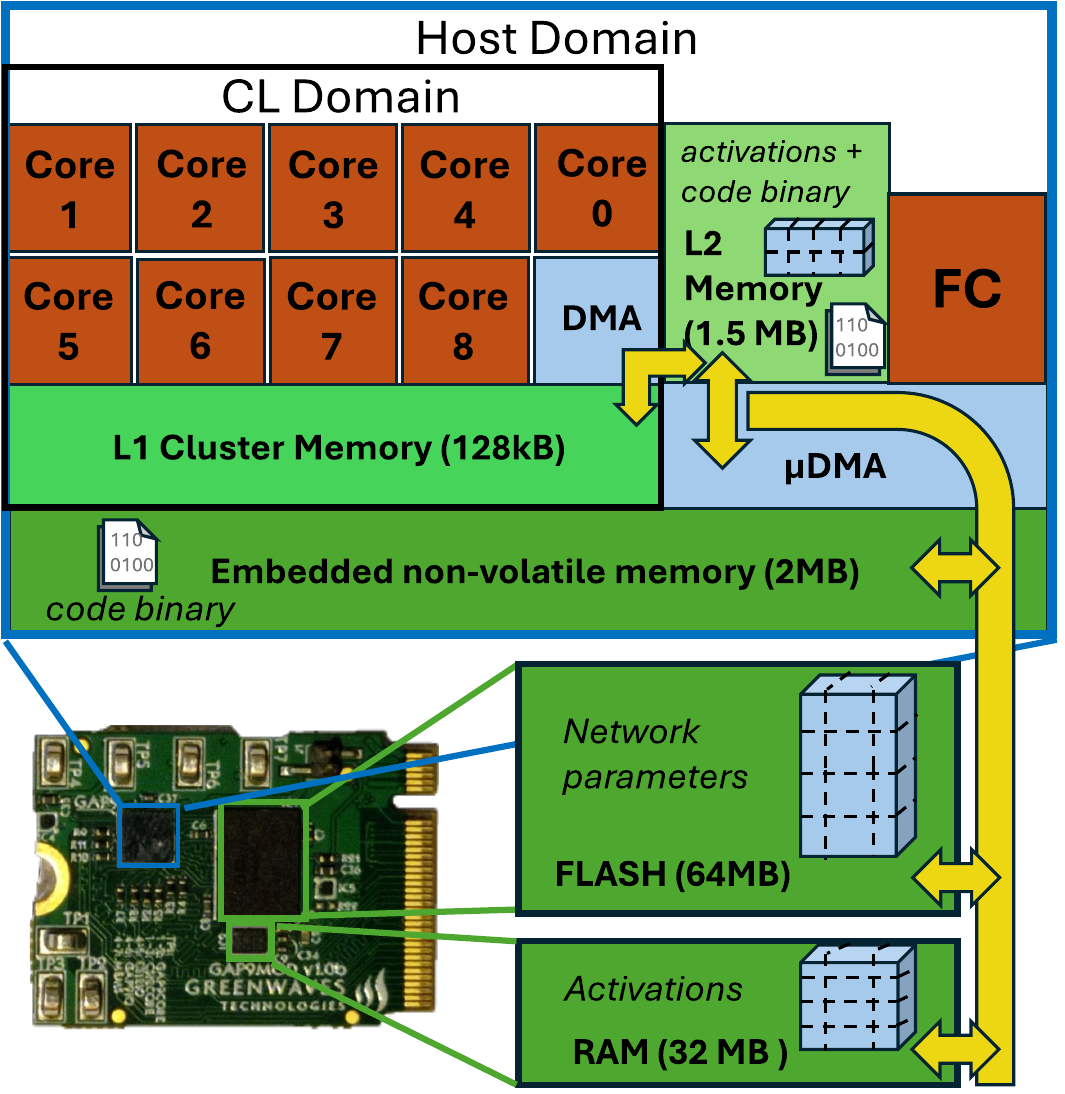}
\caption{Overview of the GAP9 MCU and the memory hierarchy. We explicitly indicate where to store the activation and weight tensors of the object detection models, as well as the binary code.}
\label{fig:gap_structure}
\end{figure}

We evaluate MR2-ByteTrack on the GAP9 MCU, chosen due to its widespread adoption in prior ultra-low-power vision systems, including autonomous drone navigation modules~\cite{Drone_movement}, smart-glass object detection platforms~\cite{TinyYOLOv2}, and compact agricultural pest monitoring nodes~\cite{pest_detection, Palossi_pest}.
As illustrated in Figure~\ref{fig:gap_structure}, GAP9 is organized into two power domains: a host domain and a compute cluster (CL).

The host domain features a single RISC-V fabric-controller (FC) core responsible for peripheral management, memory coordination, and task scheduling. It integrates \SI{1.5}{\mega\byte} of L2 memory for intermediate activations, \SI{2}{\mega\byte} of embedded non-volatile memory, and a $\mu$DMA engine that enables high-throughput transfers over an octa-SPI interface to external memory devices, which in our setup include a \SI{32}{\mega\byte} RAM and a \SI{64}{\mega\byte} Flash module.
The cluster domain serves as the primary compute accelerator. It incorporates nine RISC-V cores that support vectorized MAC operations, complemented by four shared FP16-capable floating-point units that accelerate dense linear algebra kernels. A \SI{128}{\kilo\byte} L1 scratchpad memory provides low-latency data access, whereas accesses to the L2 memory are approximately an order of magnitude slower. To mitigate this bottleneck, a dedicated DMA engine manages asynchronous L1–L2 transfers, reducing processor idle time and sustaining high computational throughput during memory-bound phases.

{{\color{blue}}Our method is not tied to a specific hardware platform and can be deployed on MCUs with comparable memory and compute capabilities, for example, the STM32N6x7 (\SI{4.2}{\mega\byte} Static Random-Access Memory (SRAM) with \SI{128}{\kilo \byte} Tightly Coupled Memory (TCM)) and NXP i.MX RT1170 (\SI{2}{\mega\byte} SRAM and \SI{512}{\kilo \byte} TCM) provides more on-chip memory than GAP9 and supports external memory, making it suitable for running the full MR2-ByteTrack pipeline. To validate portability, we also tested the most computationally demanding component, NanoDet, with full-res input ($320\times320$ px), on the STM32N6x7. More specifically, we deployed our models using the ST Edge AI Core 4.0.0 toolchain via the ST Edge AI Developer Cloud, which takes the graph-level description of the network, e.g., an Open Neural Network Exchange (ONNX) graph as input, applies calibration-based int8 post-training quantization via a per-tensor affine mapping scheme, and generates optimized C code for the target platform. The quantized model is then executed on Neural-ART, the on-chip hardware accelerator of the STM32N6x7 microcontroller. The NanoDet model with a $320\times320$ input resolution runs in \SI{98.24}{\milli\second} ms on the STM32N6570-DK evaluation board, confirming the feasibility of extending our approach beyond GAP9 to other MCU-class systems.}

To deploy CNN- or Transformer-based object detection models on the target platform, we utilize a deployment tool that transforms the ONNX graph, into low-level code. 
For the GAP9 MCU, we utilize GWT's GAP\textit{flow} toolset, which generates optimized C code for the detection task, also known as the DNN inference task. 
The tool generates header files containing model parameters and source files that define graph and layer-level routines for each trained model. 
Model weights are stored in external OctaSPI Flash memory. 
At the same time, intermediate values of the inference task, i.e., activation tensors, are allocated in the on-chip L2 memory and offloaded to external RAM when needed (see Figure~\ref {fig:gap_structure}). 
The datatype of weights and activations is cast to half-precision \texttt{FP16} for a lossless deployment. 

Because the code generation process is optimized for a given input resolution, we generate two \texttt{C} source codes to realize our multi-resolution scheme. 
Each full-resolution or low-resolution model has its own implementation, i.e., a different binary code, while both share the same weight parameters stored in Flash memory. 
A high-level scheduler selects which binary code to load at each frame to execute full-res or low-res inference. 
For CNN-based detectors, the binary codes of the two models can fit the on-chip memory. 

Conversely, the EffViT-Det model has a total code size of approximately \SI{808}{\kilo\byte}, which poses a challenge for deployment on our platform. 
Under normal operation, all executable code is permanently loaded and run from GAP9’s L2 memory.
However, in this case, the model’s code leaves insufficient space for intermediate activations, which must also reside in L2 during inference.

To address this constraint, we store the detector’s code in GAP9’s embedded non-volatile memory and rely on a code overlay mechanism to bring only the necessary portions into L2 at runtime.
Specifically, we define two overlay sections in the linker script: one for the full-resolution model and one for the low-resolution model.
We assign functions to these sections using custom compiler attributes.

At inference time, the runtime scheduler preloads the required overlay into L2 before executing the selected model. In this way, mutually exclusive code segments share the same physical L2 region, and only the code needed for the current inference occupies L2. This mechanism enables us to deploy the Transformer detector without increasing the code’s L2 memory footprint, meeting the platform’s tight memory constraints. 

%% file: 6-Results.tex
\section{Experimental Results}

\begin{table}[t]
\caption{Baseline object detection models.}
\def\arraystretch{1.2}
\centering
\resizebox{\columnwidth}{!}{%
\begin{tabular}{lcccc}
\hline
\multirow{2}{*}{\textbf{Model}} & \multirow{2}{*}{\textbf{Type}} &\multirow{2}{*}{\textbf{Parameters [\SI{}{\mega\nothing}]}} & \multicolumn{2}{c}{\textbf{MAC [\SI{}{\mega\nothing}]}} \\
&  & & \textbf{full-res} & \textbf{low-res} \\
\hline
YOLOX & CNN & 0.9 & 316 & 114 \\
NanoDet & CNN & 1.2 & 463 & 167 \\
EffViT-Det & Transformer & 0.9 & 281 & 101 \\
\hline 
\end{tabular}
}

\label{tab:Baseline}
\end{table}

\subsection{Setup}\label{sec:baseline}

We evaluate the effectiveness of MR2-ByteTrack in combination with two SoA convolution-based detectors, namely YOLOX and NanoDet, and our transformer-based detector EffVIT-Det.
Table~\ref{tab:Baseline} shows the parameter counts and computational complexity (measured in \SI{}{\mega MAC}) for processing full-res (320$\times$320 px) and low-res (192$\times$192 px) input frames.
All the detectors are trained on the COCO dataset. 
For NanoDet\footnote{Shufflenetv2-NanoDet-Plus:\url{https://github.com/RangiLyu/nanodet}} and YOLOX\footnote{ CSPDarknet-YOLOX-Nano:\url{https://github.com/Megvii-BaseDetection/YOLOX}}, we use the checkpoints available from their official repositories~\cite{ge2021yolox,nanodet}.
Instead, EffVIT-Det is trained following the training recipe proposed in~\cite{ge2021yolox}.
Specifically, the model is optimized using stochastic gradient descent with a cosine learning rate schedule and a 15-epoch warm-up phase.
The initial learning rate is 0.01.
Data augmentation techniques, like mosaic augmentation~\cite{Mosaic}, are applied during the training process and disabled for the final 15 epochs.

For our VOD experiments, we test the MR2-ByteTrack algorithms on the ImageNetVID dataset. 
We consider only video sequences containing objects from one of the 16 classes shared with the COCO dataset. 
After this filtering operation, we obtain a test subset of 392 video samples that we denote as ImageNetVID$^C$ to distinguish it from the original dataset.
{{\color{blue}} Beyond ImageNetVID$^C$, MR2-ByteTrack can be directly applied to any domain where a pretrained detector is available, since both the tracking and rescoring components are training-free and introduce no dataset-specific parameters.
ByteTrack requires that object motion remains sufficiently smooth across consecutive frames for IoU-based association to be reliable. At the same time, Rescore relies on the temporal stability of the detector's class predictions, which holds for trained models capable of generalization.
Datasets such as VisDrone~\cite{zhu2021detection}, CityFlow ~\cite{tang2019cityflow}, and UA-DETRAC~\cite{wen2020uadetrac} represent similar evaluation settings over which we expect a behavior similar to ImageNetVID$^C$.}

To assess the detection performance, we report the mAP50 metric, which we refer to as mAP throughout the rest of the paper. 
In our evaluation scheme, a detection is considered correct with respect to the ground truth if its IoU is greater than 0.5.
Additionally, we measure per-class precision, recall, and F1 scores and report the average values. 
Compared to the mAP score, these metrics are more relevant for practical deployment because they provide direct insights into the balance between true detections, false positives, and missed detections.

More specifically, for computing the metrics, we consider detections with a confidence score exceeding a threshold that maximizes the F1 score at full resolution.
The same logic is adopted for MR2-ByteTrack experiments to define the \textit{high\_threshold} and \textit{low\_threshold} parameters. 
This differs from our previous work~\cite{ours_conf} where the \textit{high\_threshold} is set to the threshold measured for the baseline models. 
This latter choice does not account for the correlation between \textit{high\_threshold} and \textit{low\_threshold}, resulting in a lower F1 score. 

{{\color{blue}}
We optimized the tracking thresholds for each network on the ImageNetVID$^C$ validation set. This choice is consistent with prior VOD works evaluating on the same dataset~\cite{2023yolov,seqnms,Seq-BBox,MEGA,TransvodLite}.
This resulted in \textit{high\_threshold} values of 0.45, 0.4, and 0.55, and \textit{low\_threshold} values of 0.3, 0.15, and 0.1, for NanoDet, YOLOX, and EffViT-Det, respectively. The IoU threshold is fixed at 0.3 for all models.
Finally, tracks are kept active for up to 5 frames even if they are not updated.
These parameters are also reported in Table~\ref{tab:tracking_params}. All networks were deployed using the GAP\textit{flow} toolset, version 5.19.5, and compiled with GCC for RISC-V, version 9.4.0. Power measurements were collected using Nordic’s Power Profiler Kit, while latency was measured through GAP9’s internal performance counters.The processor's clock frequency is set to \SI{370}{\mega\hertz}. }

\begin{table}[h]
\centering
\caption{Tracking hyperparameters for each detection model.}
\label{tab:tracking_params}
\resizebox{\columnwidth}{!}{%
\begin{tabular}{lccccc}
\hline
\multirow{2}{*}{\textbf{Model}} & \textbf{High} & \textbf{Low} & \textbf{IoU} & \textbf{Max Inactive} & \textbf{Min Consecutive} \\
 & \textbf{Thr.}  & \textbf{Thr.}  & \textbf{Thr.} $(\tau_{\mathrm{IoU}})$ & \textbf{Frames} $(\tau_{\mathrm{dead}})$ & \textbf{Matches} $(\tau_{\mathrm{init}})$ \\
\hline
NanoDet    & 0.45 & 0.30 & 0.30 & 5 & 2 \\
YOLOX      & 0.40 & 0.15 & 0.30 & 5 & 2 \\
EffViT-Det & 0.55 & 0.10 & 0.30 & 5 & 2 \\
\hline
\end{tabular}%
}
\end{table}

\subsection{Video Object Detection}\label{sec:video object Detection}

\begin{table}[t]
\caption{Comparison between baseline, Na\"ive-ByteTrack and MR2-ByteTrack VOD methods operating only with full resolution frames. Results obtained on ImageNetVID$^C$}
\def\arraystretch{1.4}
\resizebox{\columnwidth}{!}{%
\begin{tabular}{cccccc}
\hline
\textbf{Model} & \textbf{Method} & \textbf{mAP} & \textbf{Precision} & \textbf{Recall} & \textbf{F1 score} \\
\hline
\multirow{3}{*}{NanoDet} & Frame-by-frame & 48.4 & \textbf{74.6} & 52.4 & 61.6 \\
& Na\"ive-ByteTrack & 52.7 & 69.9 & 58.4 & 63.6 \\ 
& MR2-ByteTrack & \textbf{55.7} & 72.9 & \textbf{60.6} & \textbf{66.2} \\ 
\hline
\multirow{3}{*}{YOLOX} & Frame-by-frame & 48.5 & \textbf{69.9} & 52.4 & 59.9 \\
& Na\"ive-ByteTrack & 51.9 & 66.4 & 57.1 & 61.4 \\
& MR2-ByteTrack & \textbf{53.8} & 66.9 & \textbf{58.3} & \textbf{62.3} \\
\hline
\multirow{3}{*}{EffViT-Det} & Frame-by-frame & 48.6 & 68.8 & 53.1 & 59.9 \\ 
& Na\"ive-ByteTrack & 50.4 & 70.4 & 55.6 & 62.1 \\ 
& MR2-ByteTrack & \textbf{52.4} & \textbf{73.2} & \textbf{56.8} & \textbf{63.9} \\
\hline
\end{tabular}
}
\label{tab:algorithm_comparison}
\end{table}

Table~\ref{tab:algorithm_comparison} compares the performance of our MR2-ByteTrack against a baseline VOD pipeline that runs object detection on a \textit{frame-by-frame} basis.
We initially consider only detections over full-res frames, i.e., $P = 0$ low-res interleaved images.
To ablate the effectiveness of our Rescore algorithm, the table includes Naïve-ByteTrack, a variant of MR2-ByteTrack where the Rescore algorithm is omitted.

Overall, our results show that Na\"ive-ByteTrack improves the F1 score of the baseline detectors by an average of 2.09\%. 
The improvement comes from the higher number of correct detections (higher recall) obtained by using the tracking algorithm. 
By leveraging Rescore to correct the misclassified detections, MR2-ByteTrack further increases the F1 score by 1.83\%.

The frame-by-frame NanoDet baseline yields 16.8\% detections with an incorrectly predicted class, i.e., false positives.
With Na\"ive-ByteTrack, the number of correct detections increases by 20\%, but we also observe +19.3\% of false positives.
This effect arises from the tracking algorithm, which equally affects correctly and incorrectly classified objects. 
However, due to the higher number of accurate detections, the recall and F1 score of Naïve-ByteTrack are higher than those of the baseline. 
By incorporating the Rescore algorithm, MR2-ByteTrack reduces the number of false positives by 16.6\%, demonstrating that the proposed probabilistic logic can effectively correct misclassified detections.

Next, we compare YOLOX with the Transformer-based EffViT-Det, which features the same detection head but differs in its backbones. 
We first observe that for YOLOX, Naïve-ByteTrack introduces a 3\%  drop in precision compared to the baseline.
This effect arises because ByteTrack propagates false positives into subsequent frames.
Then, in MR2-ByteTrack, the Rescore algorithm reduces the false-positive rate, leading to a final F1-score 0.9\% higher than the na\"ive version.

Conversely, by applying  Na\"ive-ByteTrack to EffViT-Det, the precision and recall are increased by 1.6\% and 2.5\%, respectively.
Thanks to time-consistent outputs provided by Rescore (+2.8\% precision vs. Na\"ive-ByteTrack), MR2-ByteTrack achieves an F1 score of 63.9\%, which is 1.8\% higher than the naive version and also 1.6\% higher compared to the final score of the CNN-based counterpart.

\subsection{Multi-resolution VOD} \label{sec:multires}

\begin{figure}[t]
\includegraphics[width=\columnwidth]{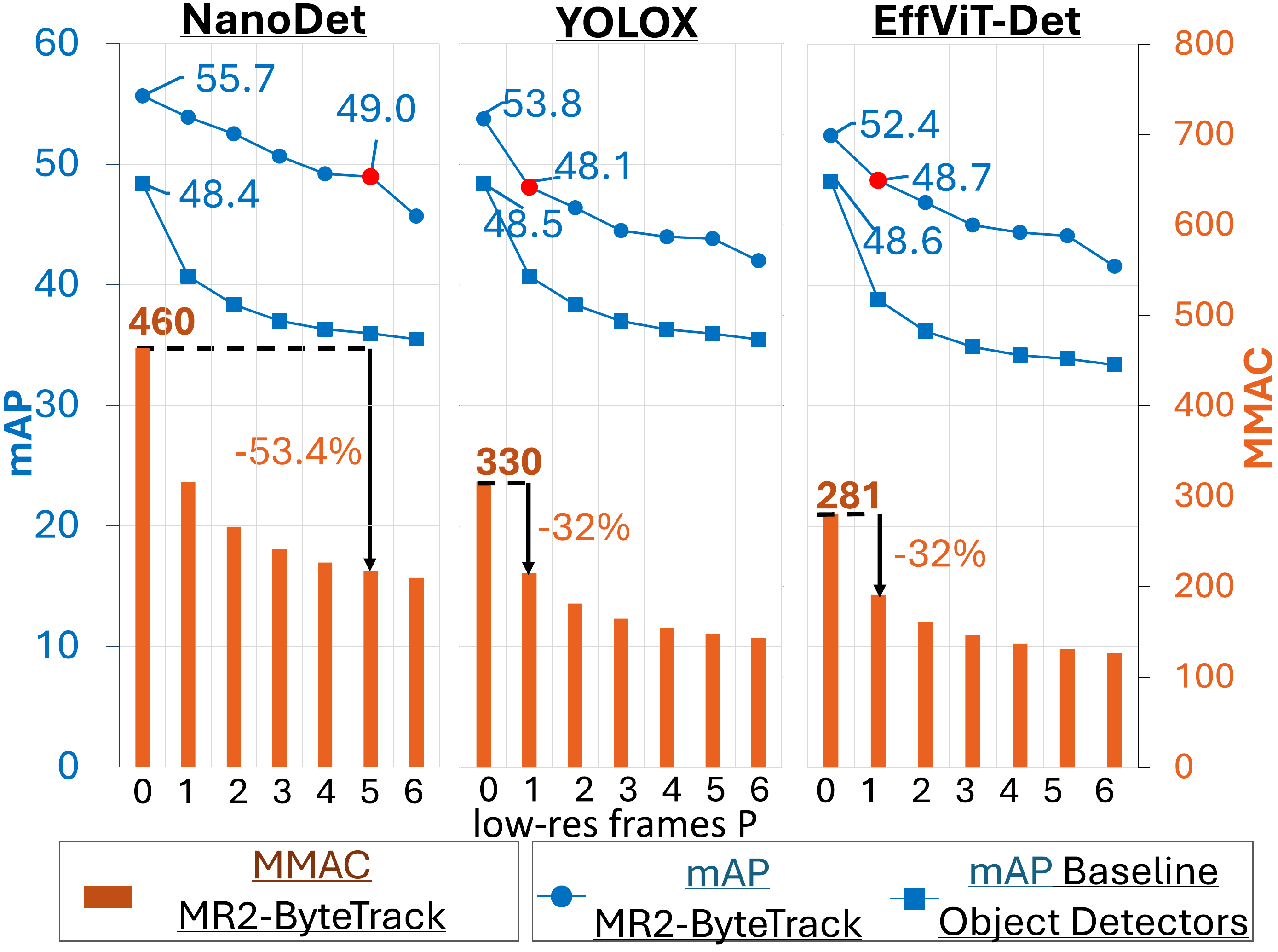}
\caption{mAP (top) and MMAC (bottom) of MR2-ByteTrack and baseline VOD solutions when varying the number of low-res frames (P) in a multi-resolution setting. In red, we highlight the MR2-ByteTrack configs with the lowest MMAC and accuracy higher or close to the baseline.}
\label{fig:result_resize}
\end{figure}

Figure~\ref{fig:result_resize} plots the mAP score of the MR2-ByteTrack models when varying the number of interleaved low-res images ($P$ from 0 to 6) in a multi-resolution setup.
For comparison purposes, we perform the same multi-resolution experiment on the frame-by-frame baseline models. 
The bars on the bottom show the \SI{}{\mega MAC} operations required by each configuration (Equation~\ref{eq:mac}).

On average, the upper plot shows an improvement of $\sim$10 mAP of MR2-ByteTrack vs. the baseline for an equal ratio of low-res and full-res images.
For the baseline models, we can observe an abrupt drop in the mAP metric already at $P = 1$. 
Compared to this, MR2-ByteTrack shows a more gradual mAP decrease with respect to the number of P frames, thanks to the tracking algorithm. 
A steep mAP drop is only observed for MR2-ByteTrack at $P = 6$. 
This effect is related to the inertia of the trackers, which remain active for five timestamps when no new detections occur.

In the upper plot of Figure~\ref{fig:result_resize}, the most computationally efficient configurations of MR2-ByteTrack are highlighted in red. 
These configurations are lossless when compared to their baseline counterparts that utilize full-resolution images.
In the case of NanoDet, at $P = 5$, we observe a 0.6\% improvement in mAP and a 53.4\% reduction in computational cost compared to the baseline. 
In contrast, YOLOX and EffViT-Det show a 32\% reduction in computational costs at $P = 1$, achieving nearly comparable mAP scores compared to the multi-resolution frame-by-frame settings (-0.4 mAP for YOLOX and +0.1 mAP for EffViT-Det).

\begin{figure}[t]
\includegraphics[width=\columnwidth]{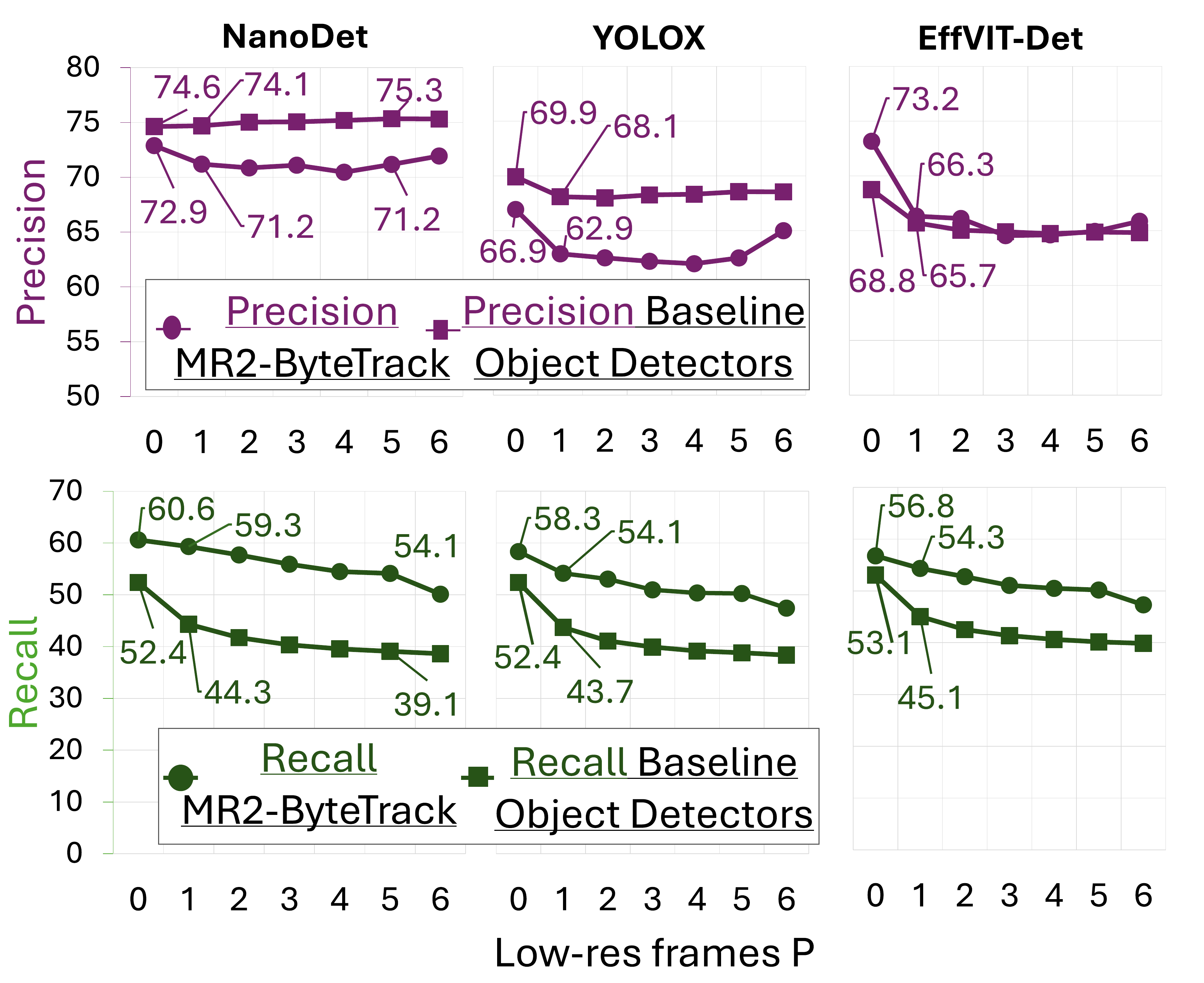}
\caption{Precision (top) and Recall (bottom) of MR2-ByteTrack and baseline VOD solutions when varying the number of low-res frames (P) in a multi-resolution
setting.}
\label{fig:result_precision_recall_resize}
\end{figure}

To provide further insights into the obtained results, Figure~\ref{fig:result_precision_recall_resize} illustrates the precision and recall scores as the number of interleaved low-resolution frames varies.  
For CNN-based object detectors (YOLOX and NanoDet), precision and recall metrics show a similar reduction when using a single low-resolution image, suggesting that the low-resolution input reduces the number of detected objects.
This behavior is arguably due to the fixed receptive fields of the convolutional layers, which limit the contextual information the model captures.
On the other side, the Transformer-based mode exhibits a different trend: while the recall remains relatively stable, precision drops significantly (Figure~\ref{fig:result_precision_recall_resize}).
We argue that this effect is due to the Transformers' ability to learn spatial relationships within an image directly. 
Hence, these models can handle objects of varying sizes and scales more effectively than CNNs, albeit with the drawback of also incurring a higher number of false positives that can lead to errors during tracking.

To further support our analysis, we examine the number of detected objects in YOLOX and EffViT-Det after integrating MR2-ByteTrack. With $P = 0$, the YOLOX network returns 20\% of false positive detections. 
The total number of detections decreases slightly at $P = 1$ by 1.4\%, while the false positive ratio remains nearly unchanged.
In contrast, EffViT-Det shows 16\% of false positives with $P = 0$, while for $P = 1$, the number of detections increases by 2.9\% and the false detection rate reaches 18.4\%. 
Hence, our analysis reveals that CNNs exhibit a reduction in precision and recall when using low-resolution images. 
In contrast, Transformers tend to produce more false detections, thereby reducing precision while maintaining a higher recall.

{{\color{blue}}
In summary, three main trends emerge from Section ~\ref{sec:video object Detection} and ~\ref{sec:multires}. 
First, ByteTrack improves recall for all three detectors.
However, for CNN-based models, this comes at the cost of a precision reduction of comparable magnitude (+6.0\% recall with -4.7\% in precision for NanoDet, +4.7\% in recall and -3.5\% in precision for YOLOX), consistent with the tracker propagating true and false positives indiscriminately.
EffViT-Det benefits from tracking across both metrics (+2.5\% recall, +1.6\% precision), suggesting that its globally-aware representations produce detections that are less sensitive to local image variations and therefore more suitable for propagation.
Second, Rescore partially mitigates this trade-off by correcting temporally inconsistent class assignments, recovering up to 3\% of the precision lost under Naïve-ByteTrack, and contributing an additional 1.83\% F1 improvement on top of the 2.09\% already provided by tracking alone. 
Third, MR2-ByteTrack exhibits a slower mAP decline as the proportion of low-resolution frames increases: while the baseline suffers an abrupt drop already at P = 1, losing 8 mAP with respect to $P = 0$,  MR2-ByteTrack maintains an average $\sim$10 mAP above the baseline at equal low-resolution ratios, with lossless operating points at P = 5 for NanoDet (+0.6 mAP, -53.4\% compute) and P = 1 for both YOLOX and EffViT-Det (-32\% compute). The degradation observed is architecture-dependent.
CNN-based detectors experience similar reductions in precision and recall, with losses of 1.7\% in precision and 1.3\% in recall for NanoDet, and losses of 4\% in precision and 4.2\% in recall for YOLOX, both evaluated at P = 1.
We argue that the reduction in precision and recall comes from the fixed receptive fields constraining multi-scale contextual capture, where EffViT-Det retains recall more effectively but incurs a higher precision reduction (6.9\% reduction in precision and 2.5\% reduction in recall), evidenced by a false positive rate rising from 16\% to 18.4\% at P = 1, as the tracker propagates false positive detections.
}

\subsection{VOD on MCUs}\label{sec:deploy}

\begin{table}[t]
\caption{Comparison between MR2-ByteTrack and frame-by-frame baselines running on the GAP9 MCU.}

\def\arraystretch{1.2}
\resizebox{\columnwidth}{!}{%
\begin{tabular}{ccccccc}
\hline
\multirow{2}{*}{\textbf{Model}} & \multirow{2}{*}{\textbf{Method}} & \textbf{RAM} & \textbf{Flash} & \textbf{Code} & \textbf{Frame-rate} & \textbf{Energy} \\
&  & \textbf{[\SI{}{\mega\byte}]} & \textbf{[\SI{}{\mega\byte}]} & \textbf{[\SI{}{\kilo\byte}]} & \textbf{[\SI{}{frame/\second}]} & \textbf{[\SI{}{\milli\joule}]} \\
\hline
\multirow{6}{*}{NanoDet} & frame-by-frame & \multirow{2}{*}{1.6} & \multirow{2}{*}{2.3} & \multirow{2}{*}{186} & \multirow{2}{*}{3.3} & \multirow{2}{*}{$21.8\pm3.3$} \\
& \textit{full-res} & & & & & \\
\cline{2-7}
& frame-by-frame & \multirow{2}{*}{0.6} & \multirow{2}{*}{2.3} & \multirow{2}{*}{186} & \multirow{2}{*}{\textbf{9.8}} & \multirow{2}{*}{$\mathbf{7.5\pm1.1}$} \\
& \textit{low-res} & & & & & \\
\cline{2-7}
& MR2-ByteTrack & \multirow{2}{*}{1.6} & \multirow{2}{*}{2.3} & \multirow{2}{*}{372} & \multirow{2}{*}{7.4} & \multirow{2}{*}{$9.9\pm1.5$} \\
& \textit{P=5} & & & & & \\
\hline
\multirow{6}{*}{YOLOX} & frame-by-frame & \multirow{2}{*}{1.4} & \multirow{2}{*}{1.8} & \multirow{2}{*}{103} & \multirow{2}{*}{3.3} & \multirow{2}{*}{$19.2\pm2.9$} \\
& \textit{full-res} & & & & & \\
\cline{2-7}
& frame-by-frame & \multirow{2}{*}{0.5} & \multirow{2}{*}{1.8} & \multirow{2}{*}{103} & \multirow{2}{*}{\textbf{8.6}} & \multirow{2}{*}{$\mathbf{7.0\pm1.1}$} \\
& \textit{low-res} & & & & & \\
\cline{2-7}
& MR2-ByteTrack & \multirow{2}{*}{1.4} & \multirow{2}{*}{1.8} & \multirow{2}{*}{206} & \multirow{2}{*}{4.8} & \multirow{2}{*}{$13.1\pm2.0$} \\
& \textit{P=1} & & & & & \\
\hline
\multirow{6}{*}{EffViT-Det} & frame-by-frame & \multirow{2}{*}{1.6} & \multirow{2}{*}{1.8} & \multirow{2}{*}{426} & \multirow{2}{*}{2.0} & \multirow{2}{*}{$25.6\pm3.8$} \\
& \textit{full-res} & & & & & \\
\cline{2-7}
& frame-by-frame & \multirow{2}{*}{0.0} & \multirow{2}{*}{1.8} & \multirow{2}{*}{382} & \multirow{2}{*}{\textbf{7.0}} & \multirow{2}{*}{$\mathbf{8.1\pm1.2}$} \\
& \textit{low-res} & & & & & \\
\cline{2-7}
& MR2-ByteTrack & \multirow{2}{*}{1.6} & \multirow{2}{*}{1.8} & \multirow{2}{*}{808} & \multirow{2}{*}{3.2} & \multirow{2}{*}{$17.1\pm2.6$} \\
& \textit{P=1} & & & & & \\
\hline
\end{tabular}
}

\label{tab:mcu_comp}
\end{table}

\begin{table*}[t]
\caption{\color{black}Comparison between VOD methods not tailored for resource-constrained MCU-based embedded systems. Results are reported on the ImagenetNetVID dataset.}
\def\arraystretch{1.2}
\centering

\begin{threeparttable}

\resizebox{\textwidth}{!}{%
\begin{tabular}{l|c|c|c|c|c|c|c|c|c|c}
\hline
\textbf{Method} &
\textbf{\shortstack{\\[0.1cm]Object detector\\Backbone}} &
\textbf{Prec.} &
\textbf{Recall} &
\textbf{\shortstack{F1\\score}} &
\textbf{mAP} &
\textbf{\shortstack{Params\\{[}M{]}}} &
\textbf{GMAC} &
\multicolumn{3}{c}{\textbf{vs. full-res object detector}} \\
\hhline{~~~~~~~~|---|}
& & & & & & & &
$\Delta$ mAP & $\Delta$ params & $\Delta$ MAC \\
\hline
YOLOV-S \cite{2023yolov} & \multirow{3}{*}{YOLOXS} & 79.9 & 66.4 & 72.5 & 62.5 & 10.3 & 12.9 & 2.7 & +14\% & +21\% \\
MR2-ByteTrack             &                         & 80.0 & 65.9 & 72.5 & 62.4 &  9.0 & 10.9 & 2.6 & 0     & 0     \\
MR2-ByteTrack ($P=2$)     &                         & 77.9 & 63.2 & 69.9 & 57.9 &  9.0 &  6.2 & -1.1& 0     & -43\% \\
\hline
\textit{Liu et. al}~\cite{Liu2019LookingFA} & SSDLite-Mobilenetv2~\cite{MobileNetv2} & n/a & n/a & n/a & 61.4 & 4.9 & 0.2 & 0.9$^1$ & +11\%$^1$ & -84\% \\
MR2-ByteTrack ($P=5$) & NanoDet$^3$~\cite{nanodet} & 71.2 & 54.1 & 61.5 & 49.0$^2$ & 1.2 & 0.2 & 0.6$^2$ & 0 & -53\% \\
\hline
\end{tabular}%
}

\begin{tablenotes}

\item[1] w.r.t. large f$_0$ non-interleaved model ~\cite{Liu2019LookingFA}. $^{2}$measured on ImageNetVID$^C$. $^{3}$trained on the COCO dataset.

\end{tablenotes}
\end{threeparttable}

\label{tab:comp}
\end{table*}


Table~\ref{tab:mcu_comp} compares our MR2-ByteTrack solution vs. frame-by-frame single-resolution baselines (e.g., used by \cite{lamberti2021low,lamberti2023bio}) with full-res and low-res inputs when running on an MCU system.
We deploy the different approaches on the GAP9 SoC, coupled with two external memories: a \SI{64}{\mega\byte} Flash and a \SI{32}{\mega\byte} RAM. 
The table reports the activation and weight memory occupation in \SI{}{\mega\byte}, denoted as RAM and Flash memory in the Table, the code size, the processing throughput in \SI{}{frame/\second}, and the energy cost, measured when the MCU is clocked at \SI{370}{\mega\hertz}.

The object detection inference tasks generally dominate the latency and memory cost of the entire MR2-ByteTrack solutions. 
{{\color{blue}} Conversely, the ByteTrack module and the Rescore algorithm incur at most \SI{0.2}{\milli\second} of execution time and \SI{11}{\kilo\byte} of memory overhead in the worst case observed during evaluation, corresponding to 21 active trackers. 
Both quantities scale linearly with the number of active trackers. Nevertheless, this overhead is negligible relative to the detector cost, remaining below $<$0.2\% of NanoDet’s inference time and below $<$1\% of its memory requirements, and is thus marginal for both CNN- and Transformer-based models.}

Regarding the object detection tasks running on the reference platform, the full-resolution model's activation memory dominates the total cost of the multi-resolution deployment. 
The inference tasks with low-res frames can reuse the same RAM space to store the intermediate results. 
Conversely, the weight storage is constant among different resolutions. 

\begin{figure}[!t]
\includegraphics[width=\columnwidth]{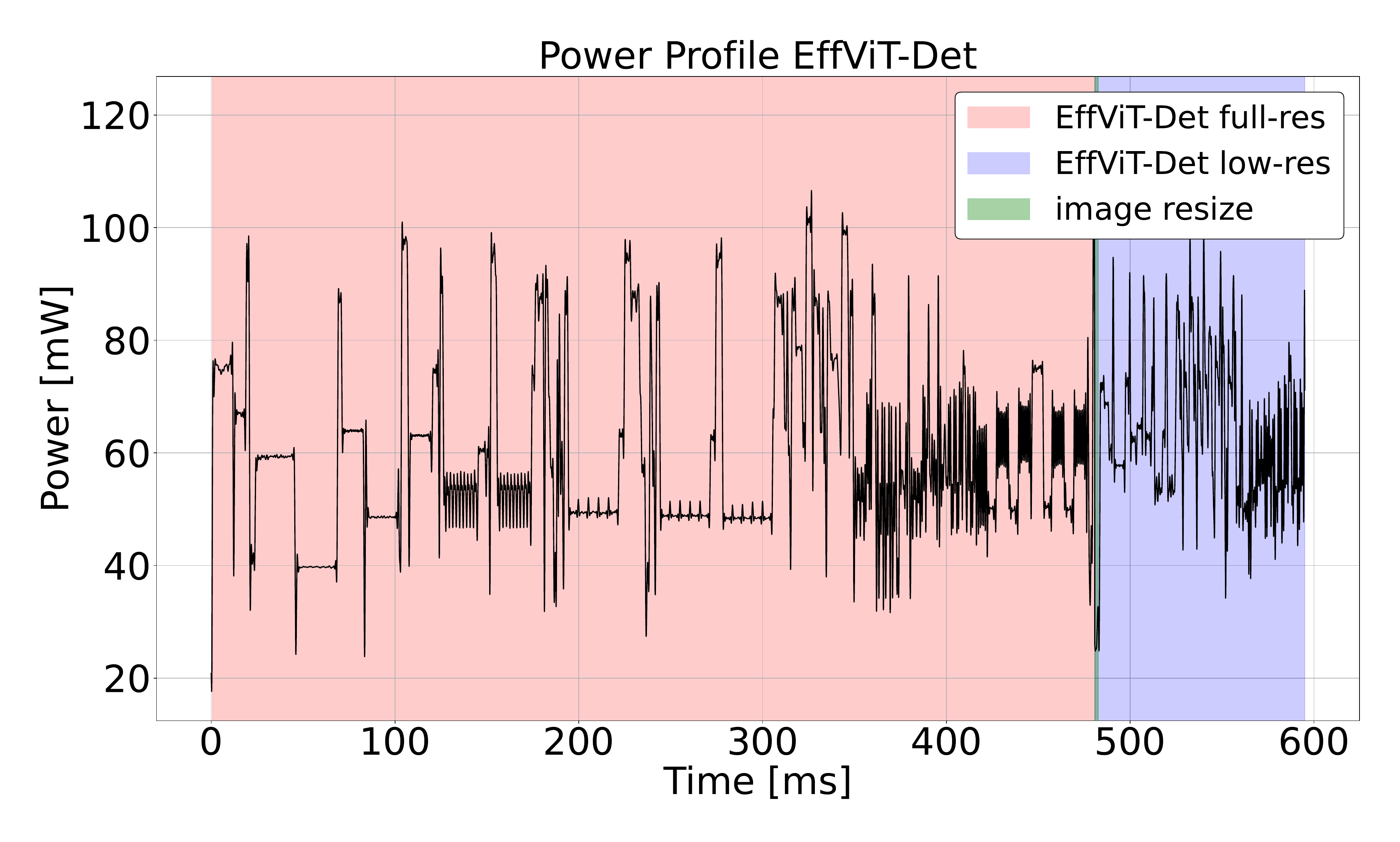}

\caption{Power profile of the EffViT-Det neural network in full-res and low-res configuration, Code loading and MR2-ByteTrack are not visible due to the limited required time.}
\label{fig:EffViTPP}
\end{figure}

While the weights and activation are retained in the external memories, the code needs to be stored in the L2 memory for fast execution. 
In the case of CNN-detectors, the memory required to store the two binary code amounts up to \SI{372}{\kilo \byte}, leaving $>$\SI{1}{\mega \byte} of free space in the L2 memory to load weight and activation data from the external memories.
Conversely, for EffViT-Det, the total code size of \SI{808}{\kilo \byte} implies only \SI{692}{\kilo \byte} of free L2 space for the intermediate results, which prevents a successful deployment.
To solve this issue, we store the double binary code in the large embedded non-volatile memory and dynamically load a resolution-specific binary into the L2 memory at runtime. 
While this copy operation introduces an overhead of \SI{0.8}{\milli\second} at an energy cost of \SI{0.023}{\milli\joule},  a total of approx ~\SI{1}{\mega\byte} of L2 memory is now available for storing network activations, which enables the deployment of EffViT-Det. The corresponding runtime power profile of the algorithm is shown in Figure~\ref{fig:EffViTPP}.
From the profile,  we can see that the execution of the detector dominates both power and energy consumption. 
In contrast, auxiliary operations, namely, image resizing for the low-resolution branch ($\sim$3 ms) and the loading of the appropriate model binary, contribute only marginally. 

Given the network model deployments, our MR2-ByteTrack solutions can run at \SI{7.4}{frame/\second} with NanoDet, \SI{4.8}{frame/\second} with YOLOX, and \SI{3.2}{frame/\second} with EffViT-Det.
The measured throughputs are, respectively, 2.2$\times$, 1.5$\times$, and 1.6$\times$ faster than the baseline full-res deployments. 
The latency gains are then reflected directly in the energy costs: the multi-resolution models consume 2.2$\times$, 1.46$\times$, and 1.5$\times$ less than the full-resolution versions for NanoDet, YOLOX, and EffViT-Det. 
Based on the reported performance scores and the accuracy results in Section~\ref{sec:multires}, \textit{our MR2-ByteTraker exhibits the best energy-accuracy trade-off among existing VOD approaches for MCU-based embedded systems.}

\subsection{VOD beyond ultra-low-power}

To showcase the generality of our approach, we apply MR2-ByteTrack to other recent real-time DNN-based methods that are \textit{not} tailored for MCU deployment because of the high number of parameters.
More in detail, we consider YOLOX-S, a CNN-based object detector featuring  \SI{9}{\mega\nothing} parameters used by the YOLOV VOD system~\cite{2023yolov}, and an SSDLite-MobileNetV2 (\SI{4.9}{\mega\nothing} parameters), which is the backbone adopted by the VOD multi-resolution strategy proposed by \textit{Liu et al.}~\cite{Liu2019LookingFA}.
The results of the comparison are reported in  Table~\ref{tab:comp}. 

To fairly compare our work with YOLOXV, we apply our MR2-ByteTrack to the same YOLOX-S backbone that was pre-trained on ImageNetVid. 
We consider a high-res or a multi-resolution configuration with P=2. 
Our solution is compared to the real-time version of YOLOV, where only the detections from past frames are sent to the transformer-based aggregator module.

Our MR2-ByteTrack achieves a +0.1\% higher precision than YOLOV, primarily due to its lower number of false positives. 
At the same time, the lower recall (-0.1\%) acknowledges the recognition ability of the final transformer layer in YOLOV, which increases the memory and computational costs by, respectively, 14\% and 21\% compared to the baseline backbones.
MR2-ByteTrack reaches the same F1 score as YOLOV but saves the transformer overhead costs.
Additionally, in the case P=2, the F1 score is reduced by only 2.6\%, but we show a computational cost reduction of up to 43\%.

On the other hand, because the pre-trained SSDLite-Mobilenetv2 model adopted by \textit{Liu et al.}~\cite{Liu2019LookingFA} is not openly available, we use the NanoDet model, which is pre-trained on COCO, for comparison (NanoDet features \SI{3.7}{\mega\nothing} parameters less than SSDLite-Mobilenetv2). 
Our MR2-ByteTrack method achieves a substantial latency improvement (-53\% vs. -84\% of \cite{Liu2019LookingFA}, which uses $P$=10), while slightly increasing the mAP score compared to the baseline.
However, our solution does not increase the memory footprint (+11\% in~\cite {Liu2019LookingFA}) and does not require training the tracker on the video sequence, highlighting the higher flexibility of our approach.

%% file: 7-Conclusion.tex
\section{Conclusions} \label{sec:conclusions}

This work investigated how on-node video processing for embedded sensory systems can be advanced under the tight compute, memory, and energy constraints of MCU-based platforms.
The proposed MR2-ByteTrack method enhances the efficiency of deployed video object detection pipelines by integrating multi-resolution inference with a lightweight mechanism that connects detections across time.
Augmenting pre-trained detectors with the ByteTrack tracker and the Rescore algorithm exploits temporal consistency, mitigating the accuracy losses introduced by low-resolution inputs. 
Experiments with memory-efficient DNNs demonstrate that MR2-ByteTrack preserves mAP while increasing throughput on the GAP9 MCU by up to 2.2$\times$.
We demonstrate the first deployment of a Transformer-based object detector on an ultra-low-power MCU, both as a standalone model and, enabled by MR2-ByteTrack, within a complete VOD pipeline, highlighting MR2-ByteTrack's capability to work with both CNN- and Transformer-based detectors.
{{{\color{blue}}}Looking forward, the tracker's bounding box predictions represent an unexploited source of spatial information that could guide a finer-grained resolution policy: rather than downsampling the entire frame uniformly, future work could selectively preserve full resolution only within tracker-predicted regions of interest, reducing compute further while concentrating representational capacity where objects are most likely to appear.
A further direction concerns the Rescore update rules, which currently treat all classes symmetrically: incorporating class-occurrence priors estimated from the detector's output statistics on unlabeled deployment data would make Rescore aware of the detector's systematic biases, for instance, down-weighting overpredicted classes, while preserving the training-free nature of the pipeline and introducing a form of lightweight unsupervised adaptation to the deployment domain.
Finally, the constant-velocity motion model underlying the current tracker represents the primary bottleneck for deployment in scenarios with irregular or ego-induced camera motion; replacing it with more complex motion priors, e.g., neural network-based trackers or extended Kalman filtering, would extend the pipeline's applicability to a broader class of platforms and use cases.
More generally, this work may pave the way for the execution of dynamic AI kernels on MCUs, enabling runtime adaptation of the compute-accuracy trade-off according to deployment-time conditions.
}

%% file: main.bib
@inproceedings{lamberti2021low,
  title={{{Low-power license plate detection and recognition on a risc-v multi-core mcu-based vision system}}},
  author={Lamberti, Lorenzo and Rusci, Manuele and Fariselli, Marco and Paci, Francesco and Benini, Luca},
  booktitle={2021 IEEE International Symposium on Circuits and Systems (ISCAS)},
  pages={1--5},
  year={2021},
  organization={IEEE}
}

@inproceedings{lamberti2023bio,
title={Bio-inspired autonomous exploration policies with cnn-based object detection on nano-drones},
author={Lamberti, Lorenzo and Bompani,Luca and Kartsch, Victor Javier and Rusci, Manuele and Palossi, Daniele and Benini, Luca},
booktitle={2023 Design, Automation \& Testin Europe Conference \& Exhibition (DATE)},
pages={1--6},
year={2023},
organization={IEEE}
}

@INPROCEEDINGS {ours_conf,
author = { Bompani, Luca and Rusci, Manuele and Palossi, Daniele and Conti, Francesco and Benini, Luca },
booktitle = { 2024 IEEE/CVF Conference on Computer Vision and Pattern Recognition Workshops (CVPRW) },
title = {{ Multi-resolution Rescored ByteTrack for Video Object Detection on Ultra-low-power Embedded Systems }},
year = {2024},
volume = {},
ISSN = {},
pages = {2182-2190},
doi = {10.1109/CVPRW63382.2024.00223},
publisher = {IEEE Computer Society},
address = {Los Alamitos, CA, USA},
month =Jun}

@inproceedings{Faster_RCNN,
author = {Ren, Shaoqing and He, Kaiming and Girshick, Ross and Sun, Jian},
title = {Faster R-CNN: towards real-time object detection with region proposal networks},
year = {2015},
publisher = {MIT Press},
address = {Cambridge, MA, USA},
booktitle = {Proceedings of the 29th International Conference on Neural Information Processing Systems - Volume 1},
pages = {91–99},
numpages = {9},
location = {Montreal, Canada},
series = {NIPS'15}
}

@inproceedings{transf_original,
author = {Vaswani, Ashish and Shazeer, Noam and Parmar, Niki and Uszkoreit, Jakob and Jones, Llion and Gomez, Aidan N. and Kaiser, \L{}ukasz and Polosukhin, Illia},
title = {Attention is all you need},
year = {2017},
isbn = {9781510860964},
publisher = {Curran Associates Inc.},
address = {Red Hook, NY, USA},
booktitle = {Proceedings of the 31st International Conference on Neural Information Processing Systems},
pages = {6000–6010},
numpages = {11},
location = {Long Beach, California, USA},
series = {NIPS'17}
}

@article{MobileNetv2,
  title={MobileNetV2: Inverted Residuals and Linear Bottlenecks},
  author={Mark Sandler and Andrew G. Howard and Menglong Zhu and Andrey Zhmoginov and Liang-Chieh Chen},
  journal={2018 IEEE/CVF Conference on Computer Vision and Pattern Recognition},
  year={2018},
  pages={4510-4520}
}

@article{Liu2019LookingFA,
  title={{{Looking fast and slow: Memory-guided mobile video object detection}}},
  author={Liu, Mason and Zhu, Menglong and White, Marie and Li, Yinxiao and Kalenichenko, Dmitry},
  journal={arXiv preprint arXiv:1903.10172},
  year={2019}
}

@inproceedings{
ViT,
title={{{An Image is Worth 16x16 Words: Transformers for Image Recognition at Scale}}},
author={Alexey Dosovitskiy and Lucas Beyer and Alexander Kolesnikov and Dirk Weissenborn and Xiaohua Zhai and Thomas Unterthiner and Mostafa Dehghani and Matthias Minderer and Georg Heigold and Sylvain Gelly and Jakob Uszkoreit and Neil Houlsby},
booktitle={International Conference on Learning Representations},
year={2021},
url={https://openreview.net/forum?id=YicbFdNTTy}
}

@unpublished{nanodet,
author={RangiLyu},
title={{{Nanodet-plus Superfast and high accuracy lightweight anchor-free object detection model}}},
url={https://github.com/RangiLyu/nanodet},
year={2021}
}

@article{ge2021yolox,
title={{{Yolox: Exceeding yolo series in 2021}}},
author={Ge, Zheng and Liu,Songtao and Wang, Feng and Li, Zeming and Sun, Jian},
journal={arXivpreprintarXiv:2107.08430},
year={2021}
}

@inproceedings{
YOLOv10,
title={{{{YOLO}v10: Real-Time End-to-End Object Detection}}},
author={Ao Wang and Hui Chen and Lihao Liu and Kai CHEN and Zijia Lin and Jungong Han and Guiguang Ding},
booktitle={The Thirty-eighth Annual Conference on Neural Information Processing Systems},
year={2024},
url={https://openreview.net/forum?id=tz83Nyb71l}
}

@misc{YOLOv11,
      title={{{YOLOv11: An Overview of the Key Architectural Enhancements}}}, 
      author={Rahima Khanam and Muhammad Hussain},
      year={2024},
      eprint={2410.17725},
      archivePrefix={arXiv},
      primaryClass={cs.CV},
      url={https://arxiv.org/abs/2410.17725}, 
}

@ARTICLE{Pass,
  author={Zhou, Qihua and Guo, Song and Pan, Jun and Liang, Jiacheng and Guo, Jingcai and Xu, Zhenda and Zhou, Jingren},
  journal={IEEE Transactions on Pattern Analysis and Machine Intelligence}, 
  title={PASS: Patch Automatic Skip Scheme for Efficient On-Device Video Perception}, 
  year={2024},
  volume={46},
  number={5},
  pages={3938-3954},
  doi={10.1109/TPAMI.2024.3350380}}

@INPROCEEDINGS{Block_copy,
  author={Verelst, Thomas and Tuytelaars, Tinne},
  booktitle={2021 IEEE/CVF International Conference on Computer Vision (ICCV)}, 
  title={{{BlockCopy: High-Resolution Video Processing with Block-Sparse Feature Propagation and Online Policies}}}, 
  year={2021},
  volume={},
  number={},
  pages={5138-5147},
  doi={10.1109/ICCV48922.2021.00511}}

@unknown{YOLOv8,
author = {Yaseen, Muhammad},
year = {2024},
month = {08},
pages = {},
title = {{{What is YOLOv8: An In-Depth Exploration of the Internal Features of the Next-Generation Object Detector}}},
doi = {10.48550/arXiv.2408.15857}
}

@inproceedings{alnuaimi2022deep,
  title={{{A Deep Learning-Based Face Mask Detector for Autonomous Nano-Drones (Student Abstract)}}},
  author={AlNuaimi, Eiman and Cereda, Elia and Psiakis, Rafail and Sugumar, Suresh and Giusti, Alessandro and Palossi, Daniele},
  booktitle={Proceedings of the AAAI Conference on Artificial Intelligence},
  volume={36},
  number={11},
  pages={12903--12904},
  year={2022}
}

@article{ILSVRC15,
Author = {Olga Russakovsky and Jia Deng and Hao Su and Jonathan Krause and Sanjeev Satheesh and Sean Ma and Zhiheng Huang and Andrej Karpathy and Aditya Khosla and Michael Bernstein and Alexander C. Berg and Li Fei-Fei},
Title = {{{ImageNet Large Scale Visual Recognition Challenge}}},
Year = {2015},
journal   = {International Journal of Computer Vision (IJCV)},
doi = {10.1007/s11263-015-0816-y},
volume={115},
number={3},
pages={211-252}
}

@InProceedings{cocodataset,
author="Lin, Tsung-Yi
and Maire, Michael
and Belongie, Serge
and Hays, James
and Perona, Pietro
and Ramanan, Deva
and Doll{\'a}r, Piotr
and Zitnick, C. Lawrence",
editor="Fleet, David
and Pajdla, Tomas
and Schiele, Bernt
and Tuytelaars, Tinne",
title="Microsoft COCO: Common Objects in Context",
booktitle="Computer Vision -- ECCV 2014",
year="2014",
publisher="Springer International Publishing",
address="Cham",
pages="740--755",
isbn="978-3-319-10602-1"
}

@InProceedings{DETR,
author="Carion, Nicolas
and Massa, Francisco
and Synnaeve, Gabriel
and Usunier, Nicolas
and Kirillov, Alexander
and Zagoruyko, Sergey",
editor="Vedaldi, Andrea
and Bischof, Horst
and Brox, Thomas
and Frahm, Jan-Michael",
title="End-to-End Object Detection with Transformers",
booktitle="Computer Vision -- ECCV 2020",
year="2020",
publisher="Springer International Publishing",
address="Cham",
pages="213--229",
isbn="978-3-030-58452-8"
}

@InProceedings{SSD,
author="Liu, Wei
and Anguelov, Dragomir
and Erhan, Dumitru
and Szegedy, Christian
and Reed, Scott
and Fu, Cheng-Yang
and Berg, Alexander C.",
editor="Leibe, Bastian
and Matas, Jiri
and Sebe, Nicu
and Welling, Max",
title={{{SSD: Single Shot MultiBox Detector}}},
booktitle="Computer Vision -- ECCV 2016",
year="2016",
publisher="Springer International Publishing",
address="Cham",
pages="21--37",
isbn="978-3-319-46448-0"
}

@inproceedings{tran2015learning,
author = {D. Tran and L. Bourdev and R. Fergus and L. Torresani and M. Paluri},
booktitle = {2015 IEEE International Conference on Computer Vision (ICCV)},
title = {{{Learning Spatiotemporal Features with 3D Convolutional Networks}}},
year = {2015},
volume = {},
issn = {2380-7504},
pages = {4489-4497},
doi = {10.1109/ICCV.2015.510},
publisher = {IEEE Computer Society},
address = {Los Alamitos, CA, USA},
month = {dec}
}

@INPROCEEDINGS{3D_V_CONV,

  author={Liu, Benzhang and Cai, Meng and Li, Jianxun},

  booktitle={2022 IEEE International Conference on Unmanned Systems (ICUS)}, 

  title={{{Video Object Detection Based on 3D Convolution}}}, 

  year={2022},

  volume={},

  number={},

  pages={177-183},

  doi={10.1109/ICUS55513.2022.9986841}}

@ARTICLE{VegaDSP,
  author={Rossi, Davide and Conti, Francesco and Eggiman, Manuel and Mauro, Alfio Di and Tagliavini, Giuseppe and Mach, Stefan and Guermandi, Marco and Pullini, Antonio and Loi, Igor and Chen, Jie and Flamand, Eric and Benini, Luca},
  journal={IEEE Journal of Solid-State Circuits}, 
  title={{{Vega: A Ten-Core SoC for IoT Endnodes With DNN Acceleration and Cognitive Wake-Up From MRAM-Based State-Retentive Sleep Mode}}}, 
  year={2022},
  volume={57},
  number={1},
  pages={127-139},
  doi={10.1109/JSSC.2021.3114881}
}

@INPROCEEDINGS{FGFA,

author={Zhu, Xizhou and Wang, Yujie and Dai, Jifeng and Yuan, Lu and Wei, Yichen},

booktitle={2017 IEEE International Conference on Computer Vision (ICCV)},

title={{{Flow-Guided Feature Aggregation for Video Object Detection}}},

year={2017},

volume={},

number={},

pages={408-417},

doi={10.1109/ICCV.2017.52}}

@article{IntegratedOD,
title={{{Integrated Object Detection and Tracking with Tracklet-Conditioned Detection}}},
author={Zheng Zhang and Dazhi Cheng and Xizhou Zhuand Stephen Lin and Jifeng Dai},
journal={ArXiv},
year={2018},
volume={abs/1811.11167},
}

@article{LYU2021139,
title={{{Video object detection with a convolutional regression tracker}}},
journal={ISPRS Journal of Photogrammetry and Remote Sensing},
volume={176},
pages={139-150},
year={2021},
issn={0924-2716},
doi={https://doi.org/10.1016/j.isprsjprs.2021.04.004},
author={Ye Lyu and Michael Ying Yang and George Vosselman and Gui-Song Xia},
}

@inproceedings{MEGA,
author={Chen, Yihong and Cao, Yue and Hu, Han and Wang, Liwei},
year={2020},
month={06},
pages={10334-10343},
title={{{Memory Enhanced Global-Local Aggregation for Video Object Detection}}},
booktitle={2020 IEEE/CVF Conference on Computer Vision and Pattern Recognition (CVPR)},
doi={10.1109/CVPR42600.2020.01035}
}

@INPROCEEDINGS{motion_based,

  author={Li, Min and Li, Linghan and Bai, Ruwen and Ren, Junxing and Meng, Bo and Yang, Yang},

  booktitle={2021 IEEE Symposium on Computers and Communications (ISCC)}, 

  title={{{A Motion-based Seq-bbox Matching Method for Video Object Detection}}}, 

  year={2021},

  volume={},

  number={},

  pages={1-7},

  doi={10.1109/ISCC53001.2021.9631435}}

@inproceedings{Seq-BBox,
title={{{Improving Video Object Detection by Seq-BboxMatching.}}},
author={Belhassen, Hatem and Zhang, Heng and Fresse, Virginie and Bourennane, El-Bay},
booktitle={VISIGRAPP(5:VISAPP)},
pages={226--233},
year={2019}
}

@article{2023yolov,
 title={{{YOLOV: Making Still Image Object Detectors Great at Video Object Detection}}}, volume={37}, DOI={10.1609/aaai.v37i2.25320}, number={2}, journal={Proceedings of the AAAI Conference on Artificial Intelligence}, author={Shi, Yuheng and Wang, Naiyan and Guo, Xiaojie}, year={2023}, month={Jun.}, pages={2254-2262} }

@inproceedings{ByteTrackMT,
  title={{{ByteTrack: Multi-Object Tracking by Associating Every Detection Box}}},
  author={Yifu Zhang and Pei Sun and Yi Jiang and Dongdong Yu and Zehuan Yuan and Ping Luo and Wenyu Liu and Xinggang Wang},
  booktitle={European Conference on Computer Vision},
  year={2021}
}

@article{seqnms,
  author = {Han, Wei and Khorrami, Pooya and Paine, Tom Le and Ramachandran, Prajit and Babaeizadeh, Mohammad and Shi, Honghui and Li, Jianan and Yan, Shuicheng and Huang, Thomas S.},
  ee = {http://arxiv.org/abs/1602.08465},
  journal = {CoRR},
  title = {{{Seq-NMS for Video Object Detection.}}},
  url = {http://dblp.uni-trier.de/db/journals/corr/corr1602.html#HanKPRBSLYH16},
  volume = {abs/1602.08465},
  year = 2016
}

@inproceedings{sort,
  author={Bewley, Alex and Ge, Zongyuan and Ott, Lionel and Ramos, Fabio and Upcroft, Ben},
  booktitle={2016 IEEE International Conference on Image Processing (ICIP)},
  title={{{Simple online and realtime tracking}}},
  year={2016},
  pages={3464-3468},
  doi={10.1109/ICIP.2016.7533003}
}

@ARTICLE{TransvodLite,
  author={Zhou, Qianyu and Li, Xiangtai and He, Lu and Yang, Yibo and Cheng, Guangliang and Tong, Yunhai and Ma, Lizhuang and Tao, Dacheng},
  journal={IEEE Transactions on Pattern Analysis and Machine Intelligence}, 
  title={{{TransVOD: End-to-End Video Object Detection With Spatial-Temporal Transformers}}}, 
  year={2023},
  volume={45},
  number={6},
  pages={7853-7869},
  doi={10.1109/TPAMI.2022.3223955}}

@INPROCEEDINGS{Streamtiny,
  author={Shalby, Hazem Hesham Yousef and Pavan, Massimo and Roveri, Manuel},
  booktitle={2024 International Joint Conference on Neural Networks (IJCNN)}, 
  title={{{StreamTinyNet: video streaming analysis with spatial-temporal TinyML}}}, 
  year={2024},
  volume={},
  number={},
  pages={1-8},
  doi={10.1109/IJCNN60899.2024.10651090}}

@INPROCEEDINGS {Objectsdonotdisappear,
author = { Liu, Xin and Nejadasl, Fatemeh Karimi and van Gemert, Jan C. and Booij, Olaf and Pintea, Silvia L. },
booktitle = { 2023 IEEE/CVF International Conference on Computer Vision (ICCV) },
title = {{{ Objects do not disappear: Video object detection by single-frame object location anticipation} }},
year = {2023},
volume = {},
ISSN = {},
pages = {6927-6938},
doi = {10.1109/ICCV51070.2023.00640},
publisher = {IEEE Computer Society},
address = {Los Alamitos, CA, USA},
month =Oct}

@article{Motetti2024AdaptiveDL,
  title={{{Adaptive Deep Learning for Efficient Visual Pose Estimation aboard Ultra-low-power Nano-drones}}},
  author={Beatrice Alessandra Motetti and Luca Crupi and Mustafa Omer Mohammed Elamin Elshaigi and Matteo Risso and Daniele Jahier Pagliari and Daniele Palossi and Alessio Burrello},
  journal={ArXiv},
  year={2024},
  volume={abs/2401.15236},
  url={https://api.semanticscholar.org/CorpusID:267312457}
}

@INPROCEEDINGS{EfficientViT,
  author={Cai, Han and Li, Junyan and Hu, Muyan and Gan, Chuang and Han, Song},
  booktitle={2023 IEEE/CVF International Conference on Computer Vision (ICCV)}, 
  title={{{EfficientViT: Lightweight Multi-Scale Attention for High-Resolution Dense Prediction}}}, 
  year={2023},
  volume={},
  number={},
  pages={17256-17267},
  doi={10.1109/ICCV51070.2023.01587}}

@article{DeformableDD,
  title={{{Deformable DETR: Deformable Transformers for End-to-End Object Detection}}},
  author={Xizhou Zhu and Weijie Su and Lewei Lu and Bin Li and Xiaogang Wang and Jifeng Dai},
  journal={ArXiv},
  year={2020},
  volume={abs/2010.04159},
  url={https://api.semanticscholar.org/CorpusID:222208633}
}

@inproceedings{
mobilevit,
title={{{MobileViT: Light-weight, General-purpose, and Mobile-friendly Vision Transformer}}},
author={Sachin Mehta and Mohammad Rastegari},
booktitle={International Conference on Learning Representations},
year={2022},
url={https://openreview.net/forum?id=vh-0sUt8HlG}
}

@article{Mosaic,
  title={YOLOv4: Optimal Speed and Accuracy of Object Detection},
  author={Alexey Bochkovskiy and Chien-Yao Wang and Hong-Yuan Mark Liao},
  journal={ArXiv},
  year={2020},
  volume={abs/2004.10934},
  url={https://api.semanticscholar.org/CorpusID:216080778}
}

@ARTICLE{Retina,
  author={Lin, Tsung-Yi and Goyal, Priya and Girshick, Ross and He, Kaiming and Dollár, Piotr},
  journal={IEEE Transactions on Pattern Analysis and Machine Intelligence}, 
  title={{{Focal Loss for Dense Object Detection}}}, 
  year={2020},
  volume={42},
  number={2},
  pages={318-327},
  doi={10.1109/TPAMI.2018.2858826}}

@misc{mobilevit3,
      title={{{MobileViTv3: Mobile-Friendly Vision Transformer with Simple and Effective Fusion of Local, Global and Input Features}}}, 
      author={Shakti N. Wadekar and Abhishek Chaurasia},
      year={2022},
      eprint={2209.15159},
      archivePrefix={arXiv},
      primaryClass={cs.CV},
      url={https://arxiv.org/abs/2209.15159}, 
}

@misc{mobilevit2,
      title={{{Separable Self-attention for Mobile Vision Transformers}}}, 
      author={Sachin Mehta and Mohammad Rastegari},
      year={2022},
      eprint={2206.02680},
      archivePrefix={arXiv},
      primaryClass={cs.CV},
      url={https://arxiv.org/abs/2206.02680}, 
}

@INPROCEEDINGS{Selsa,
  author={Wu, Haiping and Chen, Yuntao and Wang, Naiyan and Zhang, Zhao-Xiang},
  booktitle={2019 IEEE/CVF International Conference on Computer Vision (ICCV)}, 
  title={{{Sequence Level Semantics Aggregation for Video Object Detection}}}, 
  year={2019},
  volume={},
  number={},
  pages={9216-9224},
  doi={10.1109/ICCV.2019.00931}}

@inproceedings{werableAI,
  title={{{AI-based Object Detection for Assisting the Visually Impaired People}}},
  author={Sameer, Syed and Madan, Parul and Kannan, Sathish and Upadhye, Vijay Jagdish and Patil, Harshal and Rajkumar, S},
  booktitle={2024 5th International Conference on Mobile Computing and Sustainable Informatics (ICMCSI)},
  pages={512--518},
  year={2024},
  organization={IEEE}
}

@ARTICLE{Dsortmcu,
  author={Boyle, Liam and Moosmann, Julian and Baumann, Nicolas and Heo, Seonyeong and Magno, Michele},
  journal={IEEE Sensors Journal}, 
  title={{{DSORT-MCU: Detecting Small Objects in Real Time on Microcontroller Units}}}, 
  year={2024},
  volume={24},
  number={24},
  pages={40231-40239},
  doi={10.1109/JSEN.2024.3425904}}

@ARTICLE{TinyssimoYOLO,
  author={Moosmann, Julian and Müller, Hanna and Zimmerman, Nicky and Rutishauser, Georg and Benini, Luca and Magno, Michele},
  journal={IEEE Access}, 
  title={{{Flexible and Fully Quantized Lightweight TinyissimoYOLO for Ultra-Low-Power Edge Systems}}}, 
  year={2024},
  volume={12},
  number={},
  pages={75093-75107},
  doi={10.1109/ACCESS.2024.3404878}}

@ARTICLE{McuFormer,
  author={Lu, Xiwen and Bai, Chenyao and Zhu, Aoji and Zhu, Yunlong and Wang, Kezhi},
  journal={IEEE Communications Letters}, 
  title={MCFormer: A Transformer-Based Detector for Molecular Communication With Accelerated Particle-Based Solution}, 
  year={2023},
  volume={27},
  number={10},
  pages={2837-2841},
  doi={10.1109/LCOMM.2023.3303091}}

@article{Survey_CNNs_on_MCU,
author = {El Zeinaty, Christophe and Hamidouche, Wassim and Herrou, Glenn and Menard, Daniel},
title = {Designing Object Detection Models for TinyML: Foundations, Comparative Analysis, Challenges, and Emerging Solutions},
year = {2025},
issue_date = {January 2026},
publisher = {Association for Computing Machinery},
address = {New York, NY, USA},
volume = {58},
number = {2},
issn = {0360-0300},
url = {https://doi.org/10.1145/3744339},
doi = {10.1145/3744339},
journal = {ACM Comput. Surv.},
month = sep,
articleno = {50},
numpages = {48}
}

@ARTICLE{Deploy_transformers,
  author={Jung, Victor Jean-Baptiste and Burrello, Alessio and Scherer, Moritz and Conti, Francesco and Benini, Luca},
  journal={IEEE Transactions on Computers}, 
  title={{{Optimizing the Deployment of Tiny Transformers on Low-Power MCUs}}}, 
  year={2025},
  volume={74},
  number={2},
  pages={526-541},
  doi={10.1109/TC.2024.3500360}}

@INPROCEEDINGS{Deployment_trans,
  author={Burrello, Alessio and Scherer, Moritz and Zanghieri, Marcello and Conti, Francesco and Benini, Luca},
  booktitle={2021 IEEE International Conference on Omni-Layer Intelligent Systems (COINS)}, 
  title={{{A Microcontroller is All You Need: Enabling Transformer Execution on Low-Power IoT Endnodes}}}, 
  year={2021},
  volume={},
  number={},
  pages={1-6},
  doi={10.1109/COINS51742.2021.9524173}}

@Article{Transformer_mio,
AUTHOR = {Dequino, Alberto and Bompani, Luca and Benini, Luca and Conti, Francesco},
TITLE = {{{Optimizing BFloat16 Deployment of Tiny Transformers on Ultra-Low Power Extreme Edge SoCs}}},
JOURNAL = {Journal of Low Power Electronics and Applications},
VOLUME = {15},
YEAR = {2025},
NUMBER = {1},
ARTICLE-NUMBER = {8},
URL = {https://www.mdpi.com/2079-9268/15/1/8},
ISSN = {2079-9268},
DOI = {10.3390/jlpea15010008}
}

@ARTICLE{Edge_compuyting_survey,
  author={Yu, Wei and Liang, Fan and He, Xiaofei and Hatcher, William Grant and Lu, Chao and Lin, Jie and Yang, Xinyu},
  journal={IEEE Access}, 
  title={A Survey on the Edge Computing for the Internet of Things}, 
  year={2018},
  volume={6},
  number={},
  pages={6900-6919},
  doi={10.1109/ACCESS.2017.2778504}}

@article{AI_on_edge,
author = {Su, Weixing and Li, Linfeng and Liu, Fang and He, Maowei and Liang, Xiaodan},
title = {AI on the edge: a comprehensive review},
year = {2022},
issue_date = {Dec 2022},
publisher = {Kluwer Academic Publishers},
address = {USA},
volume = {55},
number = {8},
issn = {0269-2821},
url = {https://doi.org/10.1007/s10462-022-10141-4},
doi = {10.1007/s10462-022-10141-4},
journal = {Artif. Intell. Rev.},
month = dec,
pages = {6125–6183},
numpages = {59}
}

@Article{Trans_vs_cnn,
AUTHOR = {Wang, Yaoli and Deng, Yaojun and Zheng, Yuanjin and Chattopadhyay, Pratik and Wang, Lipo},
TITLE = {Vision Transformers for Image Classification: A Comparative Survey},
JOURNAL = {Technologies},
VOLUME = {13},
YEAR = {2025},
NUMBER = {1},
ARTICLE-NUMBER = {32},
URL = {https://www.mdpi.com/2227-7080/13/1/32},
ISSN = {2227-7080},
DOI = {10.3390/technologies13010032}
}

@Article{Trans_vs_cnn2,
author={Khan, Asifullah
and Rauf, Zunaira
and Sohail, Anabia
and Khan, Abdul Rehman
and Asif, Hifsa
and Asif, Aqsa
and Farooq, Umair},
title={A survey of the vision transformers and their CNN-transformer based variants},
journal={Artificial Intelligence Review},
year={2023},
month={Dec},
day={01},
volume={56},
number={3},
pages={2917-2970},
issn={1573-7462},
doi={10.1007/s10462-023-10595-0},
url={https://doi.org/10.1007/s10462-023-10595-0}
}

@ARTICLE{pest_detection,

  author={Bompani, Luca and Crupi, Luca and Palossi, Daniele and Baldoni, Olmo and Brunelli, Davide and Conti, Francesco and Rusci, Manuele and Benini, Luca},

  journal={IEEE Transactions on AgriFood Electronics}, 

  title={Accelerating Image-based Pest Detection on a Heterogeneous Multicore Microcontroller}, 

  year={2024},

  volume={2},

  number={2},

  pages={170-180},

  doi={10.1109/TAFE.2024.3451888}}

@misc{Drone_movement,
      title={A Map-free Deep Learning-based Framework for Gate-to-Gate Monocular Visual Navigation aboard Miniaturized Aerial Vehicles}, 
      author={Lorenzo Scarciglia and Antonio Paolillo and Daniele Palossi},
      year={2025},
      eprint={2503.05251},
      archivePrefix={arXiv},
      primaryClass={cs.RO},
      url={https://arxiv.org/abs/2503.05251}, 
}

@InProceedings{TinyYOLOv2,
author="Moosmann, Julian
and Bonazzi, Pietro
and Li, Yawei
and Bian, Sizhen
and Mayer, Philipp
and Benini, Luca
and Magno, Michele",
editor="Del Bue, Alessio
and Canton, Cristian
and Pont-Tuset, Jordi
and Tommasi, Tatiana",
title="Ultra-Efficient On-Device Object Detection on AI-Integrated Smart Glasses with TinyissimoYOLO",
booktitle="Computer Vision -- ECCV 2024 Workshops",
year="2025",
publisher="Springer Nature Switzerland",
address="Cham",
pages="262--280",
isbn="978-3-031-91989-3"
}

@article{Palossi_pest,
author = {Crupi, Luca and Butera, Luca and Ferrante, Alberto and Giusti, Alessandro and Palossi, Daniele},
title = {An Efficient Ground-aerial Transportation System for Pest Control Enabled by AI-based Autonomous Nano-UAVs},
year = {2025},
issue_date = {December 2025},
publisher = {Association for Computing Machinery},
address = {New York, NY, USA},
volume = {2},
number = {4},
url = {https://doi.org/10.1145/3719210},
doi = {10.1145/3719210},
journal = {ACM J. Auton. Transport. Syst.},
month = jun,
articleno = {16},
numpages = {23}
}

@ARTICLE{AI_Sensors,

  author={Mukhopadhyay, Subhas Chandra and Tyagi, Sumarga Kumar Sah and Suryadevara, Nagender Kumar and Piuri, Vincenzo and Scotti, Fabio and Zeadally, Sherali},

  journal={IEEE Sensors Journal}, 

  title={Artificial Intelligence-Based Sensors for Next Generation IoT Applications: A Review}, 

  year={2021},

  volume={21},

  number={22},

  pages={24920-24932},

  doi={10.1109/JSEN.2021.3055618}}

@ARTICLE{Agriculture,
  author={Patle, Kamlesh S and Saini, Riya and Kumar, Ahlad and Palaparthy, Vinay S.},
  journal={IEEE Sensors Journal}, 
  title={Field Evaluation of Smart Sensor System for Plant Disease Prediction Using LSTM Network}, 
  year={2022},
  volume={22},
  number={4},
  pages={3715-3725},
  doi={10.1109/JSEN.2021.3139988}}

@ARTICLE{Structural_health,

  author={Sabato, Alessandro and Dabetwar, Shweta and Kulkarni, Nitin Nagesh and Fortino, Giancarlo},

  journal={IEEE Sensors Journal}, 

  title={Noncontact Sensing Techniques for AI-Aided Structural Health Monitoring: A Systematic Review}, 

  year={2023},

  volume={23},

  number={5},

  pages={4672-4684},

  doi={10.1109/JSEN.2023.3240092}}

@article{zhu2021detection,
  title={Detection and tracking meet drones challenge},
  author={Zhu, Pengfei and Wen, Longyin and Du, Dawei and Bian, Xiao and Fan, Heng and Hu, Qinghua and Ling, Haibin},
  journal={IEEE Transactions on Pattern Analysis and Machine Intelligence},
  volume={44},
  number={11},
  pages={7380--7399},
  year={2021},
  publisher={IEEE}
}

@inproceedings{tang2019cityflow,
  author    = {Tang, Zheng and Naphade, Milind and Liu, Ming-Yu and Yang, Xiaodong and Birchfield, Stan and Wang, Shuo and Kumar, Ratnesh and Anastasiu, David and Hwang, Jenq-Neng},
  title     = {{CityFlow}: A City-Scale Benchmark for Multi-Target Multi-Camera Vehicle Tracking and Re-Identification},
  booktitle = {Proceedings of the IEEE/CVF Conference on Computer Vision and Pattern Recognition (CVPR)},
  year      = {2019}
}

@article{wen2020uadetrac,
  author    = {Wen, Longyin and Du, Dawei and Cai, Zhaowei and Lei, Zhen and Chang, Ming-Ching and Qi, Honggang and Lim, Jongwoo and Yang, Ming-Hsuan and Lyu, Siwei},
  title     = {{UA-DETRAC}: A New Benchmark and Protocol for Multi-Object Detection and Tracking},
  journal   = {Computer Vision and Image Understanding},
  volume    = {193},
  pages     = {102907},
  year      = {2020},
  doi       = {10.1016/j.cviu.2020.102907}
}
